\documentclass[runningheads]{llncs}
\usepackage{graphicx}
\usepackage{tikz}
\usepackage{comment}
\usepackage{amsmath,amssymb} % define this before the line numbering.
\usepackage{bm}

\usepackage{dsfont}
\usepackage{color}
\usepackage{enumitem}
\usepackage{caption}
\usepackage{adjustbox}
\usepackage{multirow}
\usepackage[pagebackref=true,breaklinks=true,letterpaper=true,colorlinks,bookmarks=false]{hyperref}
\usepackage{pifont}% new
\usepackage{array} % new table L/R/C
\usepackage{subcaption}
\usepackage[justification=centering]{subcaption}
\usepackage{pdfpages}

\usepackage[accsupp]{axessibility}  % Improves PDF readability for those with disabilities.

\newcommand{\ie}{\textit{i}.\textit{e}.}
\newcommand{\eg}{\textit{e}.\textit{g}.}
\newcommand{\cmark}{\ding{51}}%
\newcolumntype{L}[1]{>{\raggedright\let\newline\\\arraybackslash\hspace{0pt}}m{#1}} % new
\newcolumntype{R}[1]{>{\raggedleft\let\newline\\\arraybackslash\hspace{0pt}}m{#1}} % new
\newcolumntype{C}[1]{>{\centering\let\newline\\\arraybackslash\hspace{0pt}}m{#1}} % new
\newcommand{\rb}{\rotatebox{90}}% new

\begin{document}
% \linenumbers
\pagestyle{headings}
\mainmatter
\def\ECCVSubNumber{7048}  % Insert your submission number here

\title{OIMNet\texttt{++}: Prototypical Normalization and Localization-aware Learning for Person Search} % Replace with your title

\begin{comment}
\titlerunning{ECCV-22 submission ID \ECCVSubNumber} 
\authorrunning{ECCV-22 submission ID \ECCVSubNumber} 
\author{Anonymous ECCV submission}
\institute{Paper ID \ECCVSubNumber}
\end{comment}
%******************

%\begin{comment}
\titlerunning{OIMNet\texttt{++}}
\author{Sanghoon Lee$^{1}$ \and
Youngmin Oh$^{1}$ \and Donghyeon Baek$^{1}$ \and \\Junghyup Lee$^{1}$ \and Bumsub Ham$^{1,2}$\thanks{Corresponding author.}
\\ \url{https://cvlab.yonsei.ac.kr/projects/OIMNetPlus}}

\authorrunning{S. Lee et al.}
\institute{$^{1}$Yonsei University \quad $^{2}$Korea Institute of Science and Technology~(KIST)}
\vspace{-0.3cm}

%\end{comment}
%******************
\maketitle

\vspace{-0.5cm}
\begin{abstract}
We address the task of person search, that is, localizing and re-identifying query persons from a set of raw scene images. Recent approaches are typically built upon OIMNet, a pioneer work on person search, that learns joint person representations for performing both detection and person re-identification~(reID) tasks. To obtain the representations, they extract features from pedestrian proposals, and then project them on a unit hypersphere with L2 normalization. These methods also incorporate all positive proposals, that sufficiently overlap with the ground truth, equally to learn person representations for reID. We have found that 1)~the L2 normalization without considering feature distributions degenerates the discriminative power of person representations, and 2)~positive proposals often also depict background clutter and person overlaps, which could encode noisy features to person representations. In this paper, we introduce OIMNet\texttt{++} that addresses the aforementioned limitations. To this end, we introduce a novel normalization layer, dubbed ProtoNorm, that calibrates features from pedestrian proposals, while considering a long-tail distribution of person IDs, enabling L2 normalized person representations to be discriminative. We also propose a localization-aware feature learning scheme that encourages better-aligned proposals to contribute more in learning discriminative representations. Experimental results and analysis on standard person search benchmarks demonstrate the effectiveness of OIMNet\texttt{++}. 

\end{abstract}

\section{Introduction}
\sloppy

Person search aims at jointly localizing and re-identifying a query person from a set of raw scene images~\cite{xiao2017joint,zheng2017person}. Different from person re-identification~(reID)~\cite{ye2021deep,zheng2016person}, person search incorporates pedestrian detection in a unified framework, facilitating retrieving query persons without hand-labelled~\cite{liao2015person} or auto-detected~\cite{li2014deepreid,zheng2015scalable} pedestrian bounding boxes during inference. This provides a wide range of applications, particularly where the bounding boxes are expensive to obtain, including large-scale surveillance and pedestrian analysis. Person search is extremely challenging, since it inherits problems from both pedestrian detection~(\eg, background clutter and scale variations) and person reID~(\eg, large intra-class variations). 

\begin{figure}[t]
  \captionsetup[subfigure]{aboveskip=1pt,belowskip=1pt,justification=centering}
  \begin{adjustbox}{width=0.95\columnwidth,center} % 3x urban100 71th
    \subcaptionbox{Synthetic input}
      {\includegraphics[width=0.250\textwidth, height=0.250\textwidth]{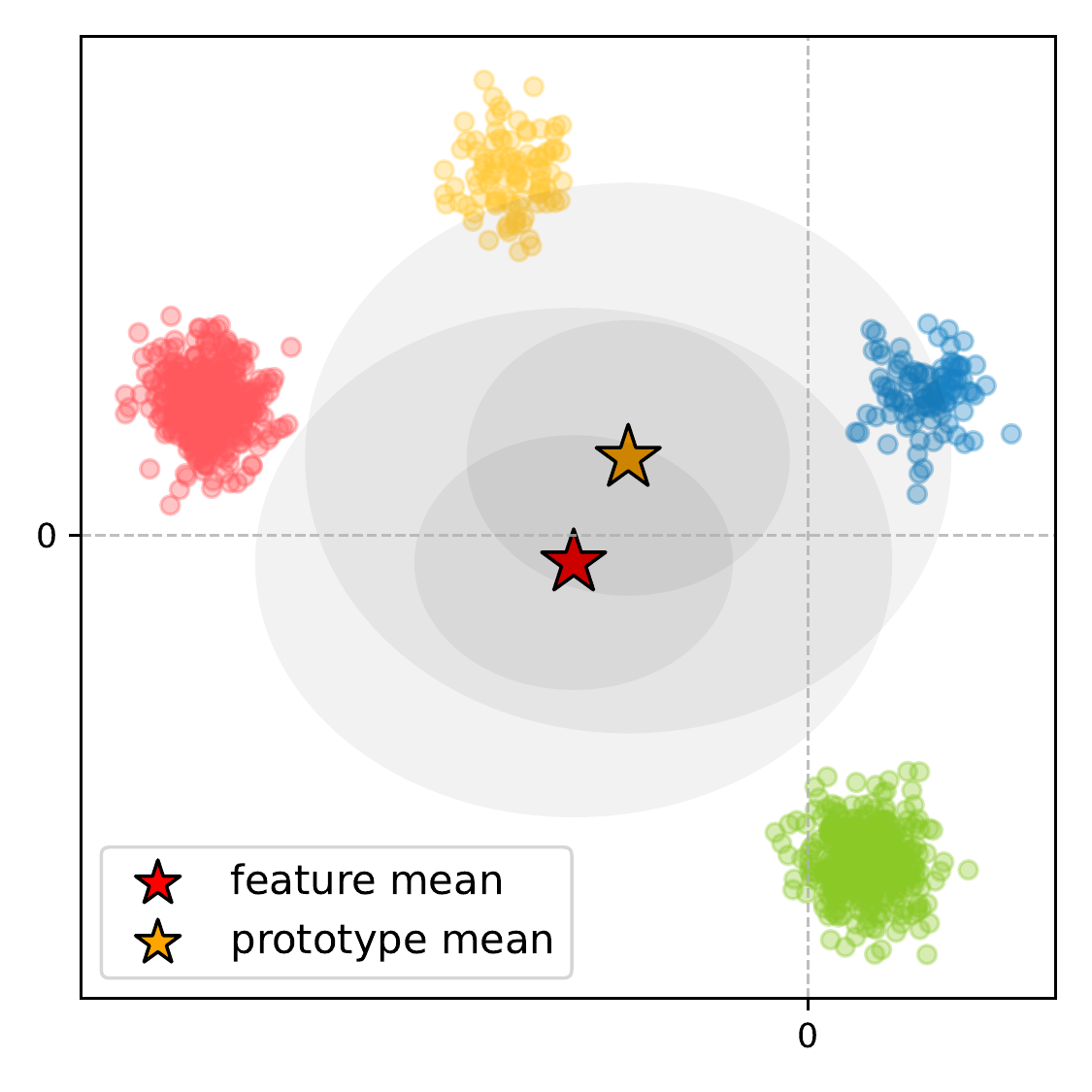}}
    \subcaptionbox{L2 Normalization}
      {\includegraphics[width=0.250\textwidth]{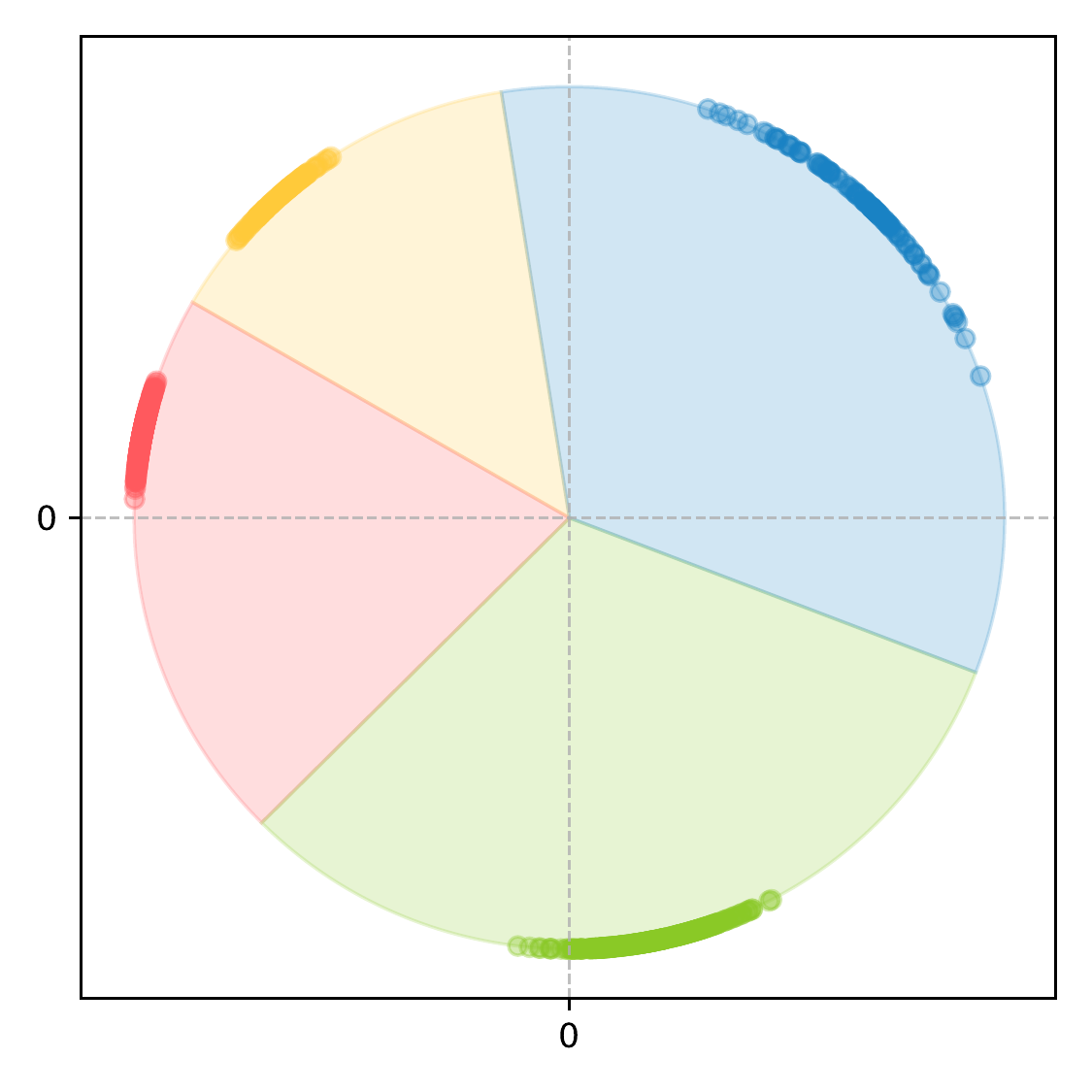}}
    \subcaptionbox{w/ BatchNorm}
     {\includegraphics[width=0.250\textwidth]{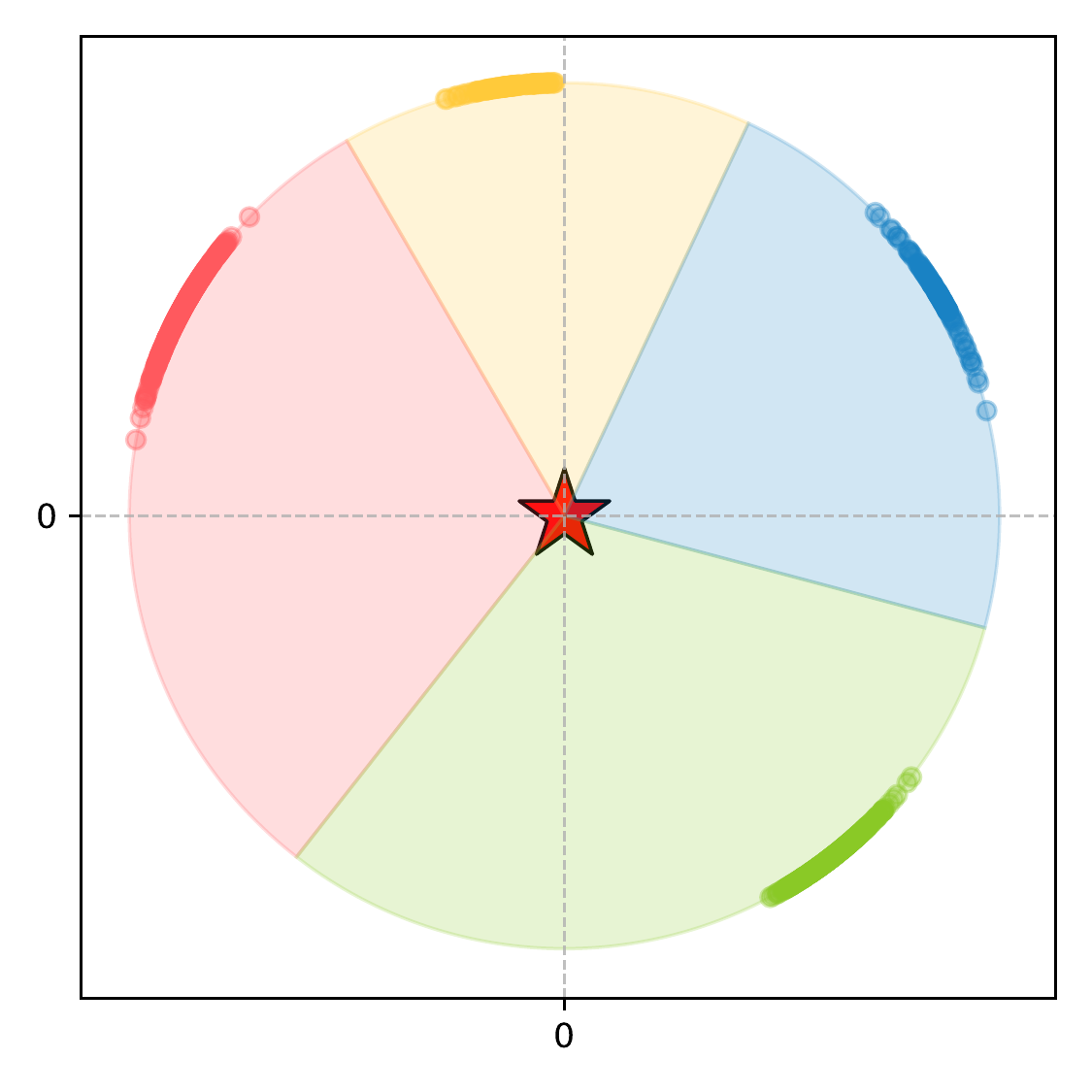}}
    \subcaptionbox{w/ ProtoNorm}
      {\includegraphics[width=0.250\textwidth]{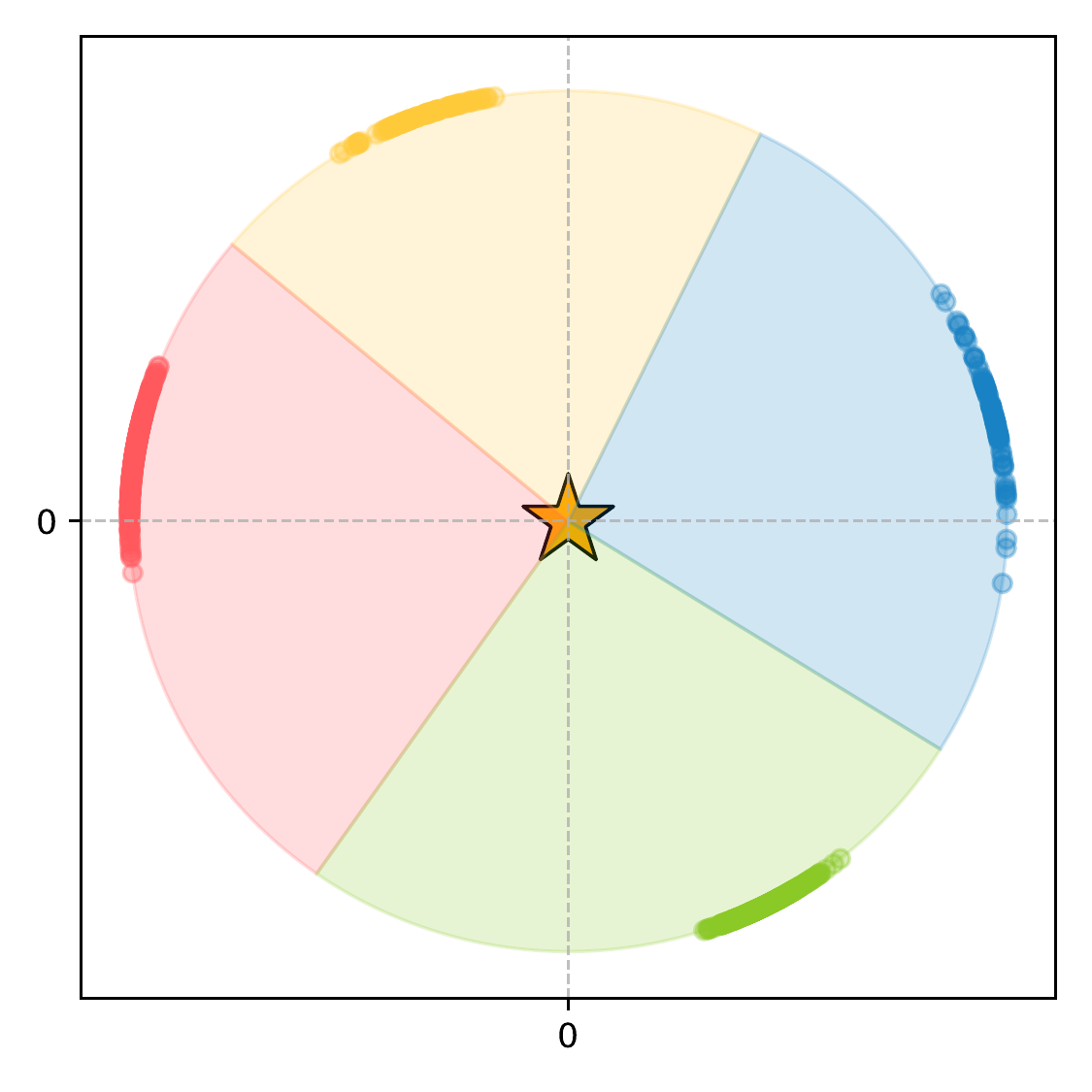}}
  \end{adjustbox}

  \captionsetup{font={small}}
  \vspace{-0.20cm}
  \caption{We visualize in (a) synthetic 2D features in circles, where each color represents an ID label. We represent mean obtained from input features and ID prototypes with stars colored in red and yellow, respectively. Note that pink and green features are sampled 4$\times$ more. The features are clearly not zero-centered with unit variance. In this case, simply applying L2 normalization degenerates the discriminative power, as shown in (b), where background colors indicate decision boundaries. Adopting a feature standardization with feature mean and variance, \ie, in a BatchNorm-fashion, prior to L2 normalization, alleviates this problem in (c). However, this does not consider a sample distribution across IDs to calibrate the feature distribution. The distribution is thus biased towards majority IDs, which weakens the inter-class separability. Instead, calibrating feature distribution using ID prototypes with ProtoNorm provides highly discriminative L2-normalized features in (d), where each ID is assigned similar angular space. (Best viewed by zooming in with color.)} 
  \label{fig:fig1}
  \vspace{-0.1cm}
\end{figure}

Recent approaches to person search focus on extracting person representations that are eligible to handle both detection and reID tasks~\cite{chen2020norm,dong2020bi,kim2021prototype,yan2021anchor}. They typically build on top of OIMNet~\cite{xiao2017joint}, a pioneer work on person search. OIMNet and its variants leverage a 2D object detection framework~\cite{ren2015faster}, and obtain person features from pedestrian proposals. The obtained feature representations are further projected onto a unit hypersphere by applying L2 normalization. These approaches also leverage the OIM loss~\cite{xiao2017joint} that employs a lookup table~(LUT) consisting of features that describe each ID in a training set, and exploits them as supervisory signals to guide learning discriminative features. While these methods have allowed significant advances for person search, there are two main limitations. First, L2 normalization is effective only when the person representations are roughly centered around zero. Moreover, the person representations are encouraged to have similar variances (\eg, unit variance) across channel dimensions over a whole training set. This is because it promotes each channel to contribute equally in determining a decision boundary, while preventing a small set of channels from dominating the decision boundary. Current person search methods implicitly assume that features obtained from pedestrian proposals have zero-mean and unit variance, \ie, standardized. Accordingly, they project the features onto a unit hypersphere using L2 normalization, without explicit constraints, which rather degenerates the discriminative power of person representations~(Fig.~\ref{fig:fig1}(b)). A standard way of alleviating this problem is to \emph{shift and scaling} the feature distribution using BatchNorm~\cite{ioffe2015batch}. However, we have observed that the standardization using BatchNorm is not able to offer satisfactory results. This is mainly due to the extreme class imbalance across person IDs, which is largely inevitable in training person search networks. Specifically, human trajectory patterns are highly diverse across person IDs, and the extent of a person's exposure to cameras in public is difficult to model~\cite{de2013unique}. Person search datasets obtained from real-world environments thus contain person images whose ID labels are extremely imbalanced, forming a long-tail distribution across person IDs. As BatchNorm calibrates the feature distribution with input features directly, it is easily biased towards dominant IDs. This restrains the discriminative power of features on a unit hypersphere~(Fig.~\ref{fig:fig1}(c)). Second, the OIM loss updates each feature in the LUT using an exponential moving average with a fixed momentum. Each object proposal contributes equally when updating the corresponding feature in the LUT, regardless of the localization accuracy w.r.t the ground truth. This is suboptimal in that not all object proposals are equally discriminative, as they are often misaligned during training. Namely, the OIM loss ignores the fact that reID relies on detection during training; better localized object proposals could contribute more to learning discriminative features for reID.

In this paper, we present a simple but highly effective approach to person search, dubbed OIMNet\texttt{++}, built upon OIMNet~\cite{xiao2017joint}. We introduce two modifications of OIMNet that offer significant performance gains over the vanilla one. First, we propose a novel normalization layer, ProtoNorm, to obtain discriminative person representations on a hypersphere. ProtoNorm exploits prototypical features for individual person IDs to calibrate feature distributions, such that the distributions are less biased towards dominant IDs~(Fig.~\ref{fig:fig1}(d)). This enhances the intra-class compactness for majority IDs, as well as the inter-class separability for minority ones. Second, we propose a localization-aware OIM loss~(LOIM) that adaptively updates each feature in the LUT w.r.t the localization accuracy of each object proposal during training. As better localized object bounding boxes translate to less noisy features, \eg, with less background clutter and person overlaps, we encourage better-aligned proposals to incorporate more to construct the LUT. In this way, we can train a network with more discriminative features as guidance. By employing ProtoNorm and the LOIM loss to a vanilla OIMNet~\cite{xiao2017joint}, we establish OIMNet\texttt{++}, setting a new state of the art on standard benchmarks for person search, including CUHK-SYSU~\cite{xiao2017joint} and PRW~\cite{zheng2017person}. The main contributions of our work can be summarized as follows:
\begin{itemize}
	\item[$\bullet$] We introduce a simple yet effective normalization layer, ProtoNorm, to learn discriminative person representations for person search. ProtoNorm alleviates the class imbalance problem in person search, while retaining the effectiveness of BatchNorm.
	\item[$\bullet$] We propose a novel LOIM loss that assigns larger degree of update to better aligned object proposals. This allows to compose the LUT with less noisy features, better guiding the discriminative feature learning for person search, compared with the OIM loss~\cite{xiao2017joint}.
	\item[$\bullet$] We set a new state of the art on standard person search benchmarks~\cite{xiao2017joint,zheng2017person} and demonstrate the effectiveness of our approach with extensive experiments and ablation studies. 
\end{itemize}

\section{Related Work}
\subsection{Person Search}

Many approaches attempt to decompose person search into pedestrian detection and person reID tasks. Current person search methods can be categorized into two groups. The first line of works~\cite{chen2018person,dong2020instance,han2019re,lan2018person} design a two-step method that performs pedestrian detection to obtain cropped person images, which are then sequentially fed into a person reID network to extract person representations. These methods further employ auxiliary modules to enhance the discriminative power of person representations. For example, they exploit an off-the-shelf instance segmentation network~\cite{li2017fully} to focus more on foreground regions~\cite{chen2018person}, design a multi-layer feature aggregation module~\cite{lan2018person}, or take a query image as an additional input with a siamese network~\cite{dong2020instance}. While these methods have achieved remarkable performances for person search, they require a heavy computational cost, due to the separate feature extractors, and prohibit joint optimization between detection and reID tasks. Another line of works~\cite{chen2020norm,dong2020bi,kim2021prototype,yan2021anchor} formulate person search as a joint feature learning problem, and optimize person search models with multi-task objectives in an end-to-end manner. Given an input image, they extract joint features for detection and reID tasks, enabling an efficient pipeline. A main challenge in training these models is known to be the contradictory objectives between detection and reID; pedestrian detection tries to extract the commonness across different person IDs, while reID focuses on the uniqueness~\cite{chen2018person,chen2020norm}. To address this problem, recent works propose to disentangle person representations into detection- and reID-related features~\cite{chen2020norm}, stop gradient flows in certain layers of a network~\cite{kim2021prototype}, or extract reID features prior to detection~\cite{yan2021anchor}. Similar to ours, the work of~\cite{kim2021prototype} proposes to adaptively update features in the LUT, while considering the hardest negative example. This, however, still ignores the localization accuracy, when adjusting momentum values for the updates. All the aforementioned person search approaches also ignore the class imbalance problem across person IDs and the corresponding detrimental effect in learning discriminative person representations.

\vspace{-0.1cm}
\subsection{Feature Normalization}
The seminal work of~\cite{ioffe2015batch} introduces a BatchNorm layer, where intermediate features of deep networks are standardized, followed by applying an affine transform. This improves the generalization capability of a network and stabilizes the training process. BatchNorm has become an indispensable component of modern deep neural networks~\cite{he2016deep,szegedy2016rethinking}. While many variants have been proposed for various  applications~\cite{ba2016layer,ioffe2017batch,li2020attentive,shao2019ssn,ulyanov2016instance,wang2018batch,wu2018group,yao2021cross}, there are no attempts to leverage normalization methods for person search, to our knowledge. Thus, we mainly describe representative works in the context of person reID, which is closely related to person search.

The work of~\cite{luo2019bag} has shown that adding a BatchNorm layer right before a final classifier boosts performance by a large margin, especially when computing distances with L2 normalized person representations. As will be shown in our experiments, this simple normalization scheme also notably improves the performance of person search methods, establishing a strong baseline. However, we have found that a vanilla BatchNorm might be susceptible to the presence of class imbalance, since statistics are dominated by majority IDs. To address this issue, ProtoNorm incorporates minority IDs to calibrate feature distributions. This standardizes features better than BatchNorm, outperforming the strong baseline. Recently, the works of~\cite{choi2021meta,jin2020style,zhou2019omni} propose to exploit both InstanceNorm~\cite{ulyanov2016instance} and BatchNorm in order to alleviate the influence of identity-irrelevant features. Although these methods achieve a better generalization ability, especially in a cross-domain setting, they still ignore the class imbalance across person IDs, when calibrating feature distributions. This suggests that our ProtoNorm further boosts the performance in a complementary way, providing more accurate feature distributions for InstanceNorm and BatchNorm. Instead of combining different normalization layers, the work of~\cite{zhuang2020rethinking} adopts multiple BatchNorm layers. It computes the mini-batch statistics separately for each camera to reduce the distribution gap across different cameras. Instead of standardizing features for each ID, we exploit a single set of statistics based on prototypical features for individual person IDs for standardization. 
\vspace{-0.2cm}

\section{Method}
\begin{figure}[t] % teaser image
  \captionsetup{font={small}}
  \begin{center}
      \includegraphics[width=1.0\columnwidth]{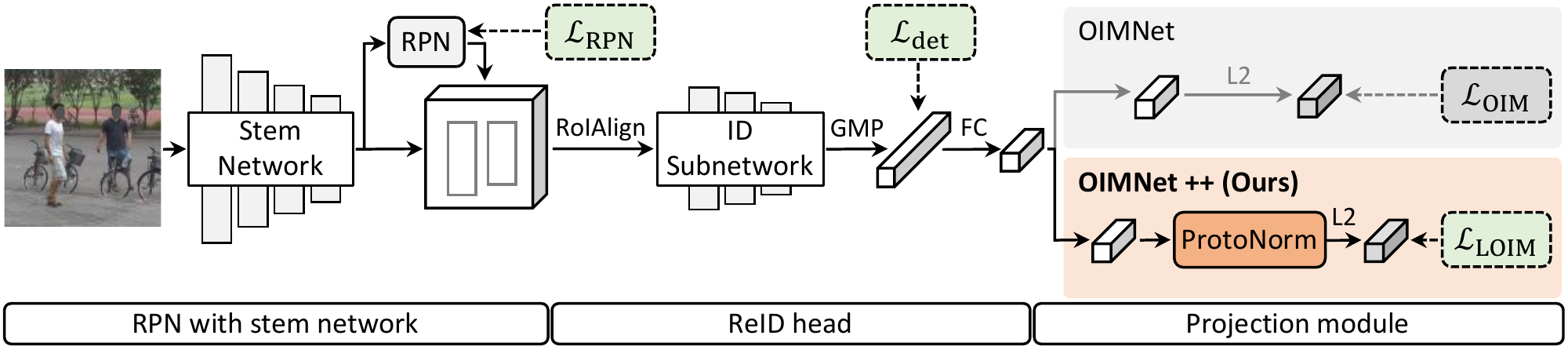}
  \end{center}
  \vspace{-0.65cm}
      \caption{
      An overview of OIMNet\texttt{++}. Similar to OIMNet~\cite{xiao2017joint}, OIMNet\texttt{++} mainly consists of three parts: An RPN with a stem network, a reID head, and a projection module. The main differences between OIMNet\texttt{++}~(bottom) and OIMNet~(top) are the projection module and the training loss. We incorporate a ProtoNorm layer to explicitly standardize features prior to L2 normalization, while considering the class imbalance problem in person search. We also exploit the LOIM loss that leverages localization accuracies of object proposals to learn discriminative features. See text for details. 
      }
  \label{fig:overview}
  \vspace{-0.2cm}
\end{figure}

We show in Fig.~\ref{fig:overview} an overview of our approach. Following the previous works~\cite{chen2020norm,dong2020bi,kim2021prototype,zhong2020robust}, we build OIMNet\texttt{++} upon OIMNet~\cite{xiao2017joint}. In this section, we briefly review OIMNet~(Sec.~3.1), and provide a detailed description of OIMNet\texttt{++}~(Sec.~3.2), including ProtoNorm and the LOIM loss. 
\vspace{-0.2cm}

\subsection{OIMNet}

\noindent\textbf{Network.} OIMNet~\cite{xiao2017joint} mainly consists of three components: A region proposal network~(RPN) with a stem network, a reID head, and a projection module. Given an input image, it employs the RPN to generate pedestrian proposals with feature maps obtained from the stem network. These proposals form candidates to be matched with a query person. While features obtained from a stem network are able to discriminate between persons and background, they are not able to discriminate between person IDs. Thus, OIMNet further employs a reID head, consisting of an identification subnetwork with a global max pooling~(GMP) layer, followed by a fully-connected~(FC) layer, to refine the features from object proposals. The projection module further projects the features on a unit hypersphere with L2 normalization. At test time, OIMNet computes distances between L2-normalized features for query and gallery persons for matching. \\

\noindent\textbf{Loss.} OIMNet is trained with the following loss:
\begin{equation}
	\mathcal{L} = 
	\lambda_{\text{RPN}}\mathcal{L}_{\text{RPN}} + 
	\lambda_{\text{det}}\mathcal{L}_{\text{det}} +
	\mathcal{L}_{\text{OIM}},
\end{equation}
where $\mathcal{L}_{\text{RPN}}$, $\mathcal{L}_{\text{det}}$, and $\mathcal{L}_{\text{OIM}}$ are RPN, detection, and OIM losses, respectively, and $\lambda_{\text{RPN}}$ and $\lambda_{\text{det}}$ are balancing parameters for corresponding terms. The RPN and detection losses consist of binary classification and offset regression terms, for anchors and proposals, respectively. They facilitate OIMNet to perform pedestrian detection.

On the other hand, to learn discriminative person representations for reID, OIMNet employs the OIM loss~\cite{xiao2017joint}. It leverages a LUT that stores features representing labelled IDs in a training set. We denote by $\mathbf{v}_{l} \in \mathbb{R}^{D}$, $l \in \{1, \cdots, L\}$ an L2-normalized feature within the LUT representing the $l$-th ID, where $L$ is the number of labelled IDs and $D$ is the channel dimension. Meanwhile, there are also pedestrian instances without corresponding ID labels. These instances form a set of unlabelled IDs that can be regarded as negatives for the labelled ones. The OIM loss also leverages a circular queue to store features obtained from the unlabelled IDs for training. Let us denote by $\mathbf{u}_{q} \in \mathbb{R}^{D}$, $q \in \{1,\cdots,Q\}$ an L2-normalized $q$-th feature within the queue, where $Q$ is the queue size. Given an L2-normalized feature $\mathbf{x}_{t} \in \mathbb{R}^{D}$, with the $t$-th ID label, the OIM loss is formally defined by 
\begin{equation}
	\mathcal{L}_{\text{OIM}} = \mathbb{E}_{\mathbf{x}}[-\log p_{t}],
\end{equation}
where 
\begin{equation}\label{eq:pt}
	p_{t} = \frac{\exp(\mathbf{v}_{t}^{\top}\mathbf{x}_{t}/\tau)}{
	\sum^{L}_{i=1}\exp(\mathbf{v}_{i}^{\top}\mathbf{x}_{t}/\tau) + \sum^{Q}_{j=1}\exp(\mathbf{u}_{j}^{\top}\mathbf{x}_{t}/\tau)},
\end{equation}
and $\tau$ is a temperature value. Namely, this term encourages an input feature to be embedded near to the corresponding ID feature within the LUT, while being distant from negative ones within the LUT and the circular queue. The OIM loss subsequently updates $\mathbf{v}_{t}$ with a fixed momentum $\eta$ as follows: 

\begin{equation}\label{eq:update}
	\mathbf{v}_{t} \leftarrow \eta \mathbf{v}_{t} + (1-\eta)\mathbf{x}_{t}.
\end{equation}
Note that $\mathbf{v}_{t}$ is L2 normalized after every update. The OIM loss stabilizes a training process, even with a large number of person IDs, and allows to leverage unlabelled IDs for learning discriminative person representations. 
 \\

\subsection{OIMNet\texttt{++}}

\begin{figure}[t] 
  \captionsetup{font={small}}
  \begin{center}
      \includegraphics[width=1.0\columnwidth]{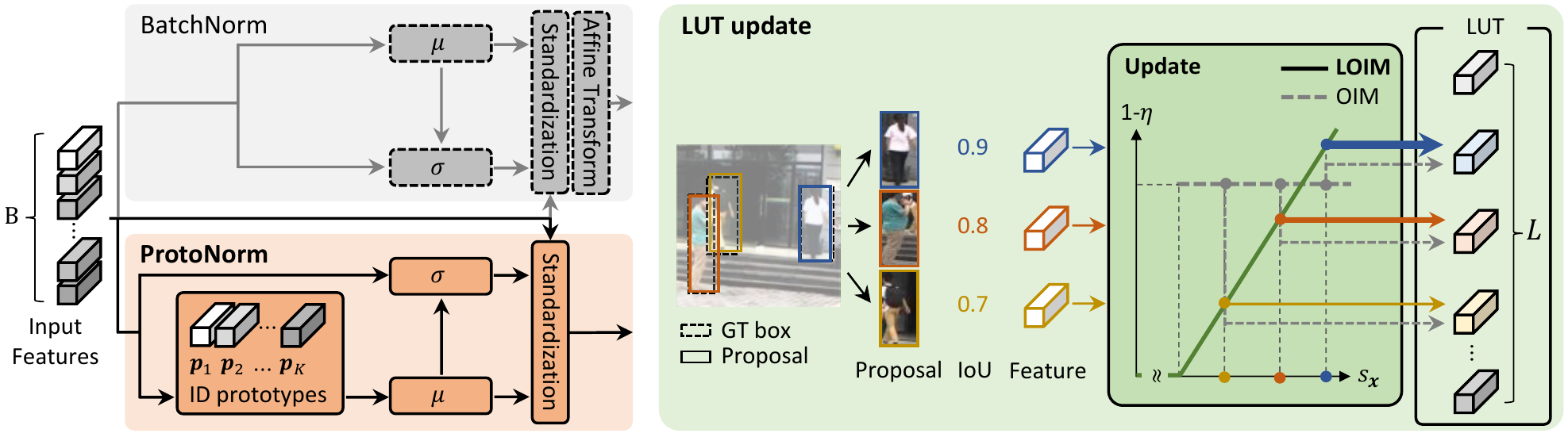}
  \end{center}
  \vspace{-0.65cm}
      \caption{\textbf{Left:} A comparison between BatchNorm~\cite{ioffe2015batch} and ProtoNorm. BatchNorm computes feature statistics with input features directly. On the other hand, ProtoNorm aggregates multiple features with the same ID into a single prototype. ProtoNorm then computes mean and variance based on the prototype features, alleviating the bias towards dominant IDs. \textbf{Right:} LUT update scheme within the  LOIM loss. The vanilla OIM loss assigns equal momentum values for all positive proposals, regardless of the localization qualities. The LOIM loss, instead, assigns an adaptive momentum value to each proposal w.r.t its IoU with the ground truth. Thicker arrows indicate larger degree of updates to the LUT. See text for details.
      }
  \label{fig:method}
  \vspace{-0.4cm}
\end{figure}

While OIMNet~\cite{xiao2017joint} has allowed significant advances for person search, there are two main limitations. First, performing L2 normalization without considering the feature distribution could be problematic, especially when features are not well standardized. In this case, L2 normalization rather degenerates the discriminative power of the features on a unit hypersphere. Note that we can employ a BatchNorm~\cite{ioffe2015batch} layer, prior to L2 normalization, that explicitly standardizes features to be zero-centered with a unit variance. As will be shown in our experiments, this improves the performance of OIMNet drastically. However, current person search datasets~\cite{xiao2017joint,zheng2017person} form a long-tail distribution across ID labels, making BatchNorm a suboptimal choice for standardization. This weakens the intra-class compactness for majority IDs and the inter-class separability for minority ones~(Fig.~\ref{fig:fig1}(c)). Second, during training, the RPN outputs pedestrian proposals that are often misaligned. On the one hand, small misalignments are acceptable, making the reID head robust to misalignment at test time. Large misalignments caused by person overlaps or mis-detections~(Fig.~\ref{fig:method}), on the other hand, distract discriminative feature learning. The OIM loss simply assumes that all proposals could contribute equally to learning discriminative features, and updates the LUT with a fixed momentum, as in (\ref{eq:update}).

In the following, we describe our approach, dubbed OIMNet\texttt{++}, that addresses the aforementioned limitations. Since OIMNet\texttt{++} and OIMNet~\cite{xiao2017joint} share the RPN and the reID head, together with the training losses for the RPN and pedestrian detection, we mainly describe ProtoNorm in the projection module and the LOIM loss in detail.\\

\noindent\textbf{ProtoNorm.} BatchNorm calibrates feature distributions using channel-wise feature statistics computed with input features directly. This makes the distributions susceptible to frequencies of ID labels, and thus they are biased towards frequent IDs. Instead of directly calibrating the distributions, we propose to exploit mini-batch statistics over prototypical features for individual person IDs. Specifically, to obtain the prototypical feature for a particular ID, we average features for the corresponding ID within a mini-batch. We provide in Fig.~\ref{fig:method} an illustration of ProtoNorm and BatchNorm.

Concretely, let us denote a set of features, $\bm{X}=\{\mathbf{x}^{1},\cdots,\mathbf{x}^{B}\}$, where $B$ is the mini-batch size, along with corresponding set of ID labels $Y=\{y^{1},\cdots,y^{B}\}$, where $y^{i} \in \{ 1,\cdots,L\}$. We denote by $\bm{X}^{i}(d)$ the $d$-th channel element within the $i$-th feature $\mathbf{x}^{i}$. We first obtain a prototypical feature representing $t$-th ID, denoted by $\mathbf{p}_{t} \in \mathbb{R}^{D}$, as follows:
\vspace{-0.15cm}
\begin{equation}
	\mathbf{p}_{t}(d) = \frac{\sum_{b=1}^{B}\bm{X}^{b}(d)\mathds{1}[y^{b}=t]}{\sum_{b=1}^{B}\mathds{1}[y^{b}=t]},
\end{equation}
where $\mathds{1}[\cdot]$ is an indicator function whose value is $1$ when the argument is true, and $0$ otherwise. With the prototypical features, $\mathbf{p}_{t}$, in hand, we compute mean and variance vectors of input features $\bm{X}$, denoted by $\bm{\mu} \in \mathbb{R}^{D}$ and $\bm{\sigma} \in \mathbb{R}^{D}$, respectively, as follows:  
\vspace{-0.15cm}
\begin{equation}\label{eq:statistic}
  \begin{split}
  	\bm{\mu}{(d)} = \frac{1}{K}  \sum_{k=1}^{K} \mathbf{p}_{k}{(d)}
  \end{split}
\quad\text{and}\quad
  \begin{split}
  	\bm{\sigma}{(d)} = \sqrt{\frac{1}{B}  \sum_{b=1}^{B} (\bm{X}^{b}(d) - \bm{\mu}{(d)})^{2}},
  \end{split}
\end{equation}
where $K$ is the number of unique IDs in the set of ID labels $Y$. We then standardize the features as $\frac{\bm{X}^{b}(d) - \bm{\mu}{(d)}}{\bm{\sigma}{(d)}}$\footnote{We could apply a learnable affine transform after standardization, similar to BatchNorm. We have empirically found that affine parameters for scaling and offset converge to constant~(but not zero) and zero values, respectively. This suggests that the effect of the affine transform is canceled out by L2 normalization, and thus we omit the transform when ProtoNorm is followed by L2 normalization.}. Instead of using input features directly, exploiting prototypical features in ProtoNorm offers standardization with mean and variance that are less biased towards dominant IDs. ProtoNorm adopts a weighted summation of the input features, where the weight can be represented as $\frac{1}{K\sum^{B}_{b=1}\mathds{1}[y^{b}=t]}$ for the $t$-th ID, and is inversely proportional to an occurrence of an ID. Namely, ProtoNorm adaptively assigns larger weight values to minority ID features, while setting to smaller ones for majority IDs. This steers the mean towards minority IDs, encouraging inter-class separateness for L2-normalized person representations. Note that we may apply the weighting method in an image level to address the class imbalance problem across person IDs,~\eg,~using a class-balanced mini-batch sampling technique~\cite{luo2019bag}. This is, however, not scalable to person search, as each image depicts different number of person instances.

Similar to BatchNorm, we track running mean and variance during training and exploit them as estimates for a global distribution of prototypical features at test time. This assumes that mean and variance sampled from the global distribution are less biased towards dominant IDs, enables calibrating the feature distribution without ID labels at test time. \\

\noindent\textbf{LOIM loss.}
The OIM loss~\cite{xiao2017joint} encourages involving positive proposals, that overlap with a ground truth more than a pre-defined threshold, equally in learning discriminative features. Since not all proposals are equally created, they should contribute to feature learning differently. The features in the LUT should thus be chosen more carefully. Specifically, the LUT should accept discriminative features only for the update, while discarding noisy ones. However, estimating the degree of noise within a feature obtained from a proposal is ambiguous. Previous works~\cite{chen2018person,zhong2020robust} have relied on an auxiliary supervision obtained from, \eg, an off-the-shelf pose estimator~\cite{fang2017rmpe}, or an instance segmentation network~\cite{li2017fully}. They are computationally expensive, and require additional datasets for training. Intersection-of-union~(IoU) between a proposal and its ground-truth bounding box, on the other hand, serves as a good indicator for estimating extent of noise within the proposal. Namely, a proposal with a large IoU score tightly covers a person-of-interest, with less background clutter and person overlaps. Note that ground-truth bounding boxes are already available in person search datasets, suggesting that leveraging IoU scores does not require additional labelling effort. To implement this idea, we exploit the IoU score of each proposal to update features in the LUT. We assign small momentum values for proposals with large IoU scores, as these proposals are able to provide less noisy features. Let us denote by $s_{\mathbf{x}} \in [0,1]$ the IoU score between a proposal and its ground-truth bounding box. Concretely, we update the features within the LUT as follows:
\begin{equation}\label{eq:newupdate}
	\mathbf{v}_{t} \leftarrow (1-c_{\mathbf{x}})\mathbf{v}_{t} + c_{\mathbf{x}}\mathbf{x}_{t},
\end{equation}
where $c_{\mathbf{x}}$ computes an adaptive momentum, defined using the IoU score, as follows:
\begin{equation}
	c_{\mathbf{x}} = \text{clip}(s_{\mathbf{x}}, 0, 1-\epsilon).
\end{equation}
$\text{clip}(\cdot,0,1-\epsilon)$ is a clipping function with lower and upper bounds set to $0$ and $1-\epsilon$, respectively, and $\epsilon$ is a hyperparameter. We simply set $\epsilon$ to $0.1$ to prevent perfectly-aligned proposals~(\ie,~$s_{\mathbf{x}}=1$) from totally overriding the corresponding feature after the LUT update. The LOIM loss is defined by $\mathcal{L}_{\text{LOIM}} = \mathbb{E}_{\mathbf{x}}[-\log p_{t}]$, whereas $\mathbf{v}_{t}$ is updated with the adaptive momentum, as in (\ref{eq:newupdate}). By leveraging IoU scores to define momentum values, noisy features are discouraged to form the LUT, better guiding discriminative feature learning. This also encourages a network to favor extracting better-localized features at test time.

\section{Experiments}
\subsection{Implementation details}

\noindent\textbf{Network.} Following the previous works~\cite{chen2020norm,dong2020bi,kim2021prototype,zhong2020robust}, we exploit ResNet50~\cite{he2016deep} pretrained on ImageNet~\cite{ILSVRC15} as our backbone network. Concretely, we split ResNet50 at \texttt{conv4-6} layer, and establish a stem network and an identification subnetwork with preceding and succeeding layers, respectively. We employ RoIAlign~\cite{he2017mask} to crop $14 \times 14$ proposal feature maps obtained from the stem network, and set the channel dimension of person representations, $D$, to $256$, following OIMNet~\cite{xiao2017joint}. We leverage the circular queue with size, $Q$, $5000$ and $500$ for CUHK-SYSU~\cite{xiao2017joint} and PRW~\cite{zheng2017person}, respectively. Note that we exploit feature maps obtained from \texttt{conv5-3} of ResNet50 only as inputs to the RoIAlign layer, rather than fusing multi-level features in a pyramid fashion~\cite{lin2017feature}. 
 \\

\begin{figure}[t]
  \captionsetup[subfigure]{aboveskip=1pt,belowskip=1pt,justification=centering}
  \begin{adjustbox}{width=0.8\columnwidth,center} % 3x urban100 71th
    \subcaptionbox{PRW~\cite{zheng2017person}}
      {\includegraphics[width=0.4\textwidth]{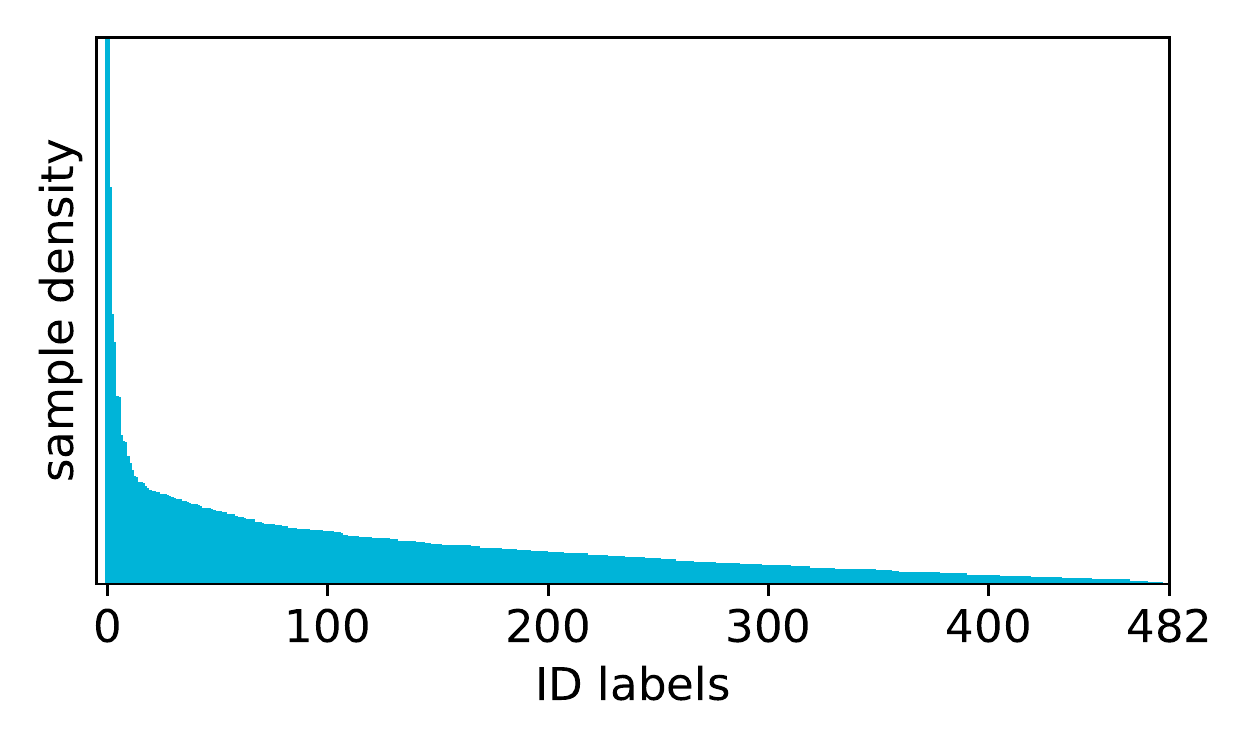}}
    \subcaptionbox{CUHK-SYSU~\cite{xiao2017joint}}
      {\includegraphics[width=0.4\textwidth]{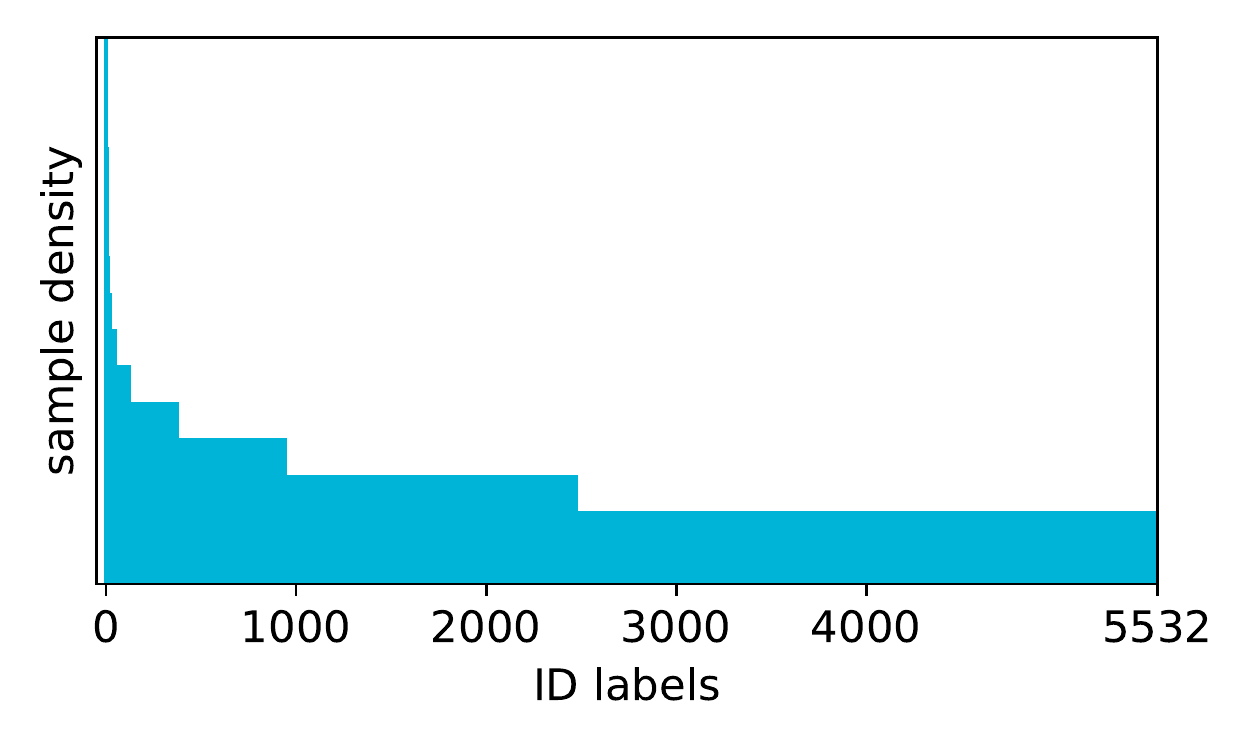}}
  \end{adjustbox}

  \captionsetup{font={small}}
  \vspace{-0.25cm}
  \caption{Density of training samples across ID labels in PRW~\cite{zheng2017person} and CUHK-SYSU~\cite{xiao2017joint} datasets. Since human trajectory patterns in public are highly diverse across persons~\cite{de2013unique}, person IDs, collected from a real-world environment, are extremely imbalanced, forming a long-tail distribution.}
  \label{fig:dataset}
  \vspace{-0.5cm}  
\end{figure}

\noindent\textbf{Dataset.} We use two standard benchmarks for training and evaluation: 1)~The PRW dataset~\cite{zheng2017person} is collected using $6$ cameras in a university. It includes $11,816$ images with $4,310$ pedestrian bounding boxes, which are labelled to $932$ IDs. We adopt the train/test splits provided by the authors, and use $5,704$ images containing $482$ IDs for training. There are $2,057$ query images with $450$ different IDs, whereas a gallery set contains $6,112$ images. For each query image, we use the whole gallery set for evaluation. 2)~The CUHK-SYSU dataset~\cite{xiao2017joint} is collected from urban scenes and movie clips. It is composed of $18,184$ images that contain $96,143$ pedestrian  bounding boxes with $8,432$ labelled IDs. We use official train/test splits provided by the authors. Concretely, we use $11,206$ images with $5,532$ IDs for training, $6,978$ gallery images with $2,900$ query person instances for testing. Following the standard protocol~\cite{xiao2017joint}, we sample $100$ gallery images for each query person during evaluation. For both datasets, we adjust input images to the size of $900 \times 1,500$ for training and testing. We visualize in Fig.~\ref{fig:dataset} the distributions for the number of training samples across ID labels. \\

\noindent\textbf{Training.} We use the same training strategy and hyperparameter setting as the ones in~\cite{chen2020norm}. Specifically, we train our model for $20$ epochs for both the PRW~\cite{zheng2017person} and CUHK-SYSU~\cite{xiao2017joint} datasets, and set the batch size to $5$. We employ a warm-up strategy, gradually increasing a learning rate to $0.003$ during the first epoch, which is divided by $10$ at the $16$th epoch. We assign the same balancing parameter for each term in the training objective, \ie, $\lambda_{\text{RPN}}=1$ and $\lambda_{\text{det}}=1$, and set the temperature value $\tau$ to $0.33$. We set the momentum value $\eta$ to $0.5$ for training a network with the OIM loss. We train our model using PyTorch~\cite{paszke2017automatic} end-to-end, which takes about $5$ and $10$ hours for PRW and CUHK-SYSU datasets, respectively, with a Titan RTX GPU. 
\vspace{-0.4cm}

\begin{table}[t]
  \captionsetup{font={small}}
  \centering
    \caption{Quantitative comparison with the state of the art for person search. We report mAP~(\%) and rank-1 accuracy~(\%) on CUHK-SYSU~\cite{xiao2017joint} and PRW~\cite{zheng2017person} datasets. For each category of person search methods, numbers in bold indicate the best and underscored ones indicate the second best. R50 and DC are abbreviations for ResNet50~\cite{he2016deep} and deformable convolution~\cite{dai2017deformable}, respectively. We report our average scores over $4$ runs with standard deviations in parentheses.}\label{table:sota}
    \begin{tabular}{C{0.4cm}|C{3.0cm}|C{2.5cm}|C{1.2cm}C{1.2cm}|C{1.2cm}C{1.2cm}}
      \hline
      & \multirow{2}{*}{Method} & \multirow{2}{*}{Backbone} & \multicolumn{2}{c|}{CUHK-SYSU~\cite{xiao2017joint}}                        & \multicolumn{2}{c}{PRW~\cite{zheng2017person}}                              \\
      &                         &                           & \multicolumn{1}{c|}{mAP} & \multicolumn{1}{c|}{rank-1} & \multicolumn{1}{c|}{mAP} & \multicolumn{1}{c}{rank-1} \\
      \hline \hline
      \multirow{4}{*}{\rb{Two-step}}    & MGTS~\cite{chen2018person}     & R50              & 83.0             & 83.7             & 32.6             & 72.1             \\
                                        & RDLR~\cite{han2019re}     & R50              & \underline{93.0} & \underline{94.2} & 42.9             & 70.2             \\
                                        & IGPN~\cite{dong2020instance}     & R50              & 90.3             & 91.4             & \bf{47.2} & \underline{87.0} \\
                                        & TCTS~\cite{wang2020tcts}     & R50              & \bf{93.9}        & \bf{95.1}        & \underline{46.8}        & \bf{87.5}        \\
      \hline\hline
      \multirow{10}{*}{\rb{End-to-end}} 
      & OIM~\cite{xiao2017joint} & R50 & 75.5 & 78.7 & 21.3 & 49.4 \\
      & NPSM~\cite{liu2017neural} & R50 & 77.9 & 81.2 & 24.2 & 53.1 \\
      & QEEPS~\cite{munjal2019query} & R50 & 88.9 & 89.1 & 37.1 & 76.7 \\
      & NAE\texttt{+}~\cite{chen2020norm} & R50 & \underline{92.1} & 92.9 & 44.0 & 81.1 \\
      & BINet~\cite{dong2020bi} & R50 & 90.0 & 90.7 & 45.3 & 81.7 \\
      & PGA~\cite{kim2021prototype} & R50 & 90.2 & 91.8 & 42.5 & \underline{83.5} \\
      & AlignPS~\cite{yan2021anchor}  & R50 & \textbf{93.1} & \underline{93.4} & \underline{45.9} & 81.9 \\
      & OIMNet\texttt{++}~(Ours) & R50 & \textbf{93.1} \tiny{(0.24)} & \textbf{93.9} \tiny{(0.30)}  & \textbf{46.8} \tiny{(0.51)} & \textbf{83.9} \tiny{(0.59)} \\ \cline{2-7}
      & PGA*~\cite{kim2021prototype} & R50-Dilation & 92.3 & \textbf{94.7} & 44.2 & \textbf{85.2} \\
      & AlignPS\texttt{+}~\cite{yan2021anchor} & R50-DC~\cite{dai2017deformable} & \textbf{94.0} & \underline{94.5} & \underline{46.1} & 82.1 \\
      & OIMNet\texttt{+++}~(Ours) & R50-ProtoNorm & \underline{93.1} \tiny{(0.21)} & 94.1 \tiny{(0.25)} & \textbf{47.7} \tiny{(0.19)} & \underline{84.8} \tiny{(0.20)}\\
      \hline\hline
  \end{tabular}
\end{table}

\subsection{Comparison with the state of the art}
We provide in Table~\ref{table:sota} a quantitative comparison between our method with the state of the art~\cite{chen2018person,chen2020norm,dong2020bi,dong2020instance,han2019re,kim2021prototype,liu2017neural,munjal2019query,wang2020tcts,xiao2017joint,yan2021anchor} for person search. For fair comparison, we categorize person search methods into two-step~\cite{chen2018person,dong2020instance,han2019re,wang2020tcts} and end-to-end~\cite{chen2020norm,dong2020bi,kim2021prototype,liu2017neural,munjal2019query,xiao2017joint,yan2021anchor} approaches. The end-to-end approaches are further split into two groups according to the backbone network. 

Overall, we can see from the experimental results that OIMNet\texttt{++} provides highly discriminative person representations for person search. In particular, OIMNet\texttt{++} shows high mAP scores. This indicates that our model is able to offer retrieval results with less false positives, \ie, matches that are not likely to be a false alarm. Among the end-to-end approaches that adopt vanilla ResNet50~\cite{he2016deep} as a backbone network, OIMNet\texttt{++} achieves the state-of-the-art performance. Note that OIMNet\texttt{++} even outperforms PGA~\cite{kim2021prototype} that requires additional parameters and computational overhead at test time due to an auxiliary attention module. 

Recent works~\cite{kim2021prototype,yan2021anchor} modify a backbone network to further boost the performance. For example, PGA*~\cite{kim2021prototype} provides a variant by reducing the dilation rate of a \texttt{conv5} block in ResNet50 from $2$ to $1$ to obtain features of high resolution. AlignPS\texttt{+}~\cite{yan2021anchor} additionally exploits deformable convolutions from \texttt{conv3} to \texttt{conv5} blocks within ResNet50. Similarly, we replace BatchNorm layers within a \texttt{conv5} block with ProtoNorm for OIMNet\texttt{+++}. In this case, we apply a learnable affine transformation after ProtoNorm layers within ResNet50, as in BatchNorm. Note that our modification, compared to other ones for PGA~\cite{kim2021prototype} and AlignPS~\cite{yan2021anchor}, does not require additional computational overheads or parameters at test time. This places our model at a disadvantage, but we can see from the the results in the last row of Table~\ref{table:sota} that OIMNet\texttt{+++} shows the person search performances comparable with competitive approaches, even including two-step ones~\cite{chen2018person,dong2020instance,han2019re,wang2020tcts}.

\begin{table}[t]
  \captionsetup{font={small}}
  \centering
  \caption{Ablative analysis of our approach. We measure the mAP~(\%) and rank-1 accuracy~(\%) on PRW~\cite{zheng2017person} using person representations obtained from detected and annotated bounding boxes to evaluate search and reID performances separately. BN and PN indicates BatchNorm~\cite{ioffe2015batch} and ProtoNorm, respectively. Numbers in bold indicate the best performance and the underscored ones indicate the second best. All results are obtained by averaging scores over $4$ runs.}\label{table:ablation}
  \begin{tabular}{C{1.2cm} C{1.2cm} C{1.2cm} C{1.2cm}|C{1.2cm} C{1.2cm}|C{1.2cm} C{1.2cm}}
    \hline
    \multirow{2}{*}{BN} & \multirow{2}{*}{PN} & \multirow{2}{*}{$\mathcal{L}_{\text{OIM}}$} & \multirow{2}{*}{$\mathcal{L}_{\text{LOIM}}$} & \multicolumn{2}{c|}{Search} & \multicolumn{2}{c}{ReID} \\
    & & & & \multicolumn{1}{c|}{mAP} & rank-1 & \multicolumn{1}{c|}{mAP} & rank-1 \\ 
    \hline \hline
           &        & \cmark &        & 42.0 & 80.5 & 44.3 & 82.6 \\
    \cmark &        & \cmark &        & 44.3 & 81.5 & 46.6 & 83.2 \\
           & \cmark & \cmark &        & \underline{46.3} & 82.7 & \underline{48.4} & 84.6 \\
    \cmark &        &        & \cmark & 45.1 & \underline{82.9} & 47.4 & \underline{84.7} \\
           & \cmark &        & \cmark & \textbf{46.8} & \textbf{83.9} & \textbf{49.0} & \textbf{86.2} \\
    \hline\hline
  \end{tabular}
  \vspace{-0.2cm}
\end{table}

\subsection{Discussion}

\noindent\textbf{Ablation study.} We provide in Table~\ref{table:ablation} an ablation study of our approach using different combinations of components and losses. We measure mAP~(\%) and rank-1 accuracy~(\%) on the test set of PRW~\cite{zheng2017person}. To better evaluate the discriminative power of person representations, we also measure the reID performance using annotated bounding boxes. We can see from the first and second rows that a person search model trained with the OIM loss only shows the worst performance\footnote{The model in the first row is exactly same as the original OIMNet~\cite{xiao2017joint}, apart from the RoIAlign module in ours. Note that re-implementing OIMNet using common practices in recent works~\cite{chen2020norm,kim2021prototype,li2021sequential}~(an improved learning rate scheduler, larger batch size, and the RoIAlign module) performs significantly better than the original OIMNet shown in Table~\ref{table:sota}. Similar findings are also reported in~\cite{chen2020norm,li2021sequential}.}, and incorporating a BatchNorm layer can boost the performance significantly. This suggests that applying the L2 normalization to person representations without standardization techniques degenerates the discriminative power. By replacing BatchNorm with our ProtoNorm in the third row, we can achieve additional performance gains. This demonstrates the effectiveness of ProtoNorm calibrating feature distributions while explicitly considering the class imbalance problem in person search. The results coincide with our finding in the toy experiment illustrated in Fig.~\ref{fig:fig1}, confirming once more the importance of a class-unbiased standardization scheme prior to projecting features on a unit hypersphere. We can observe from the second and fourth rows that the LOIM loss boosts the performance drastically. This suggests that selectively updating LUT features with the localization accuracy in the LOIM loss helps learning more discriminative representations than the OIM loss. Lastly, jointly exploiting ProtoNorm and the LOIM loss in the last row shows the best performance, and the two components complement each other. 

\begin{figure}[t]
  \captionsetup[subfigure]{aboveskip=1pt,belowskip=1pt,justification=centering}
  \begin{adjustbox}{width=0.9\columnwidth,center} % 3x urban100 71th
    \subcaptionbox{LUT}
      {\includegraphics[width=0.32\textwidth]{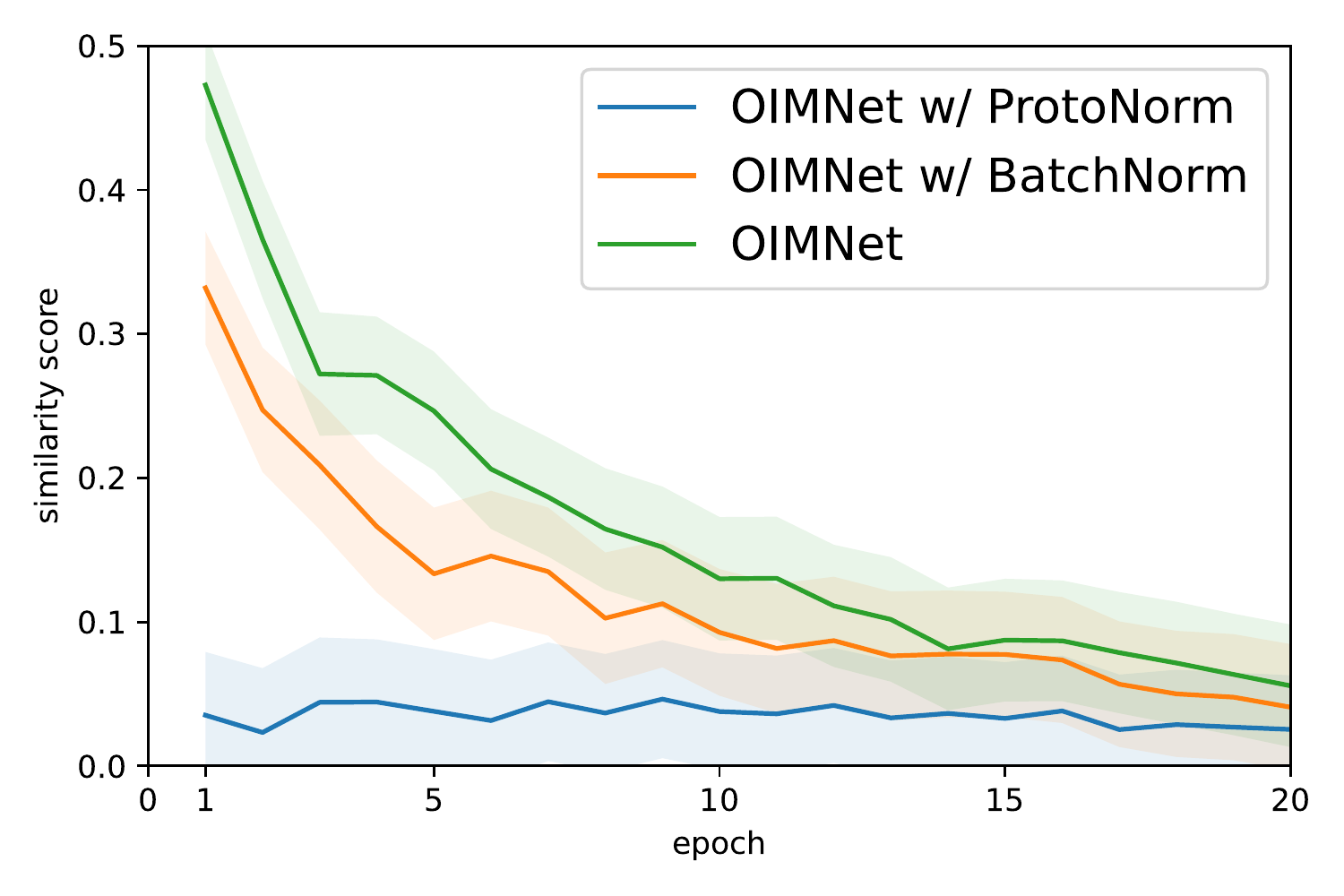}}
    \subcaptionbox{Circular queue}
      {\includegraphics[width=0.32\textwidth]{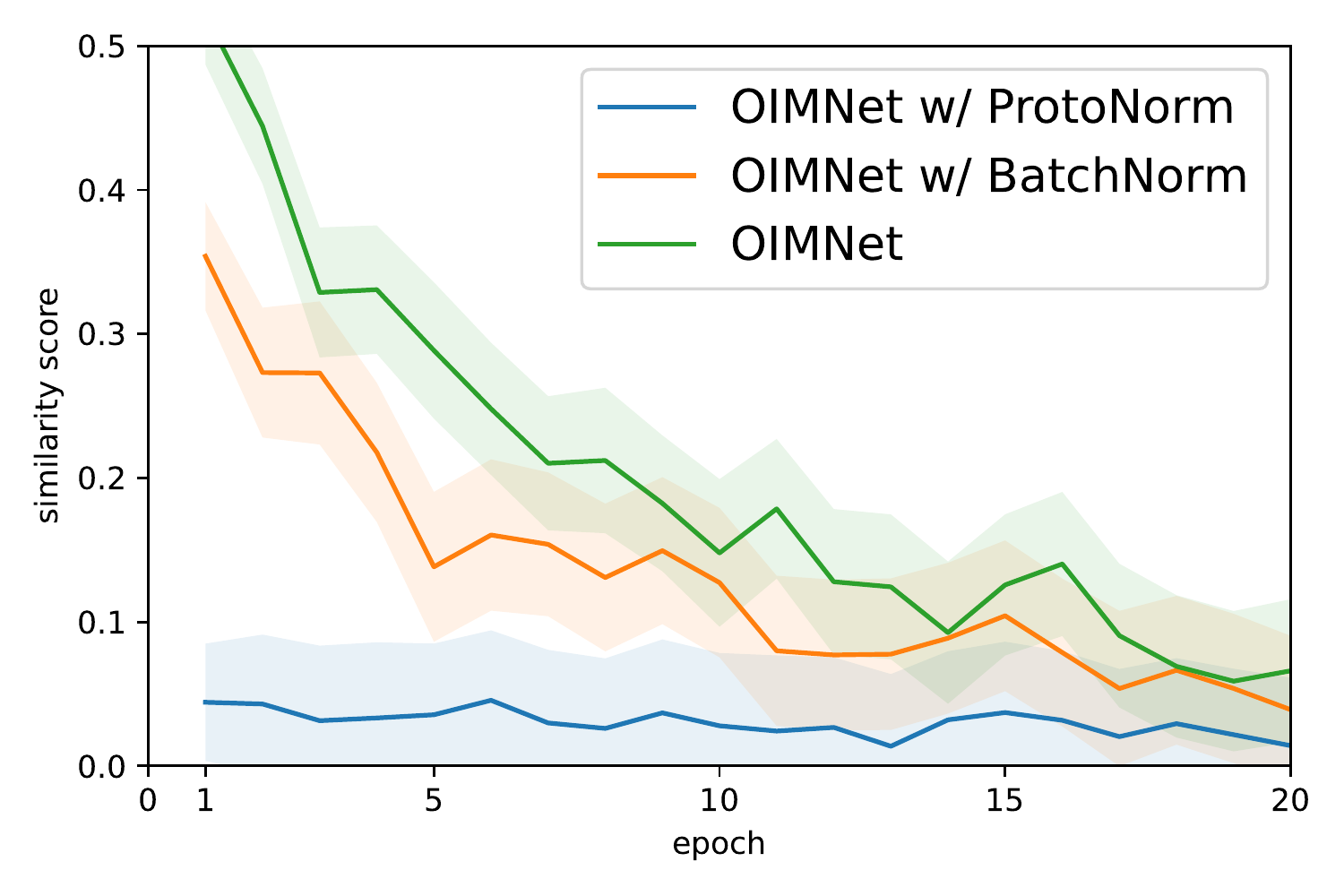}}
	    \subcaptionbox{Combined}
      {\includegraphics[width=0.32\textwidth]{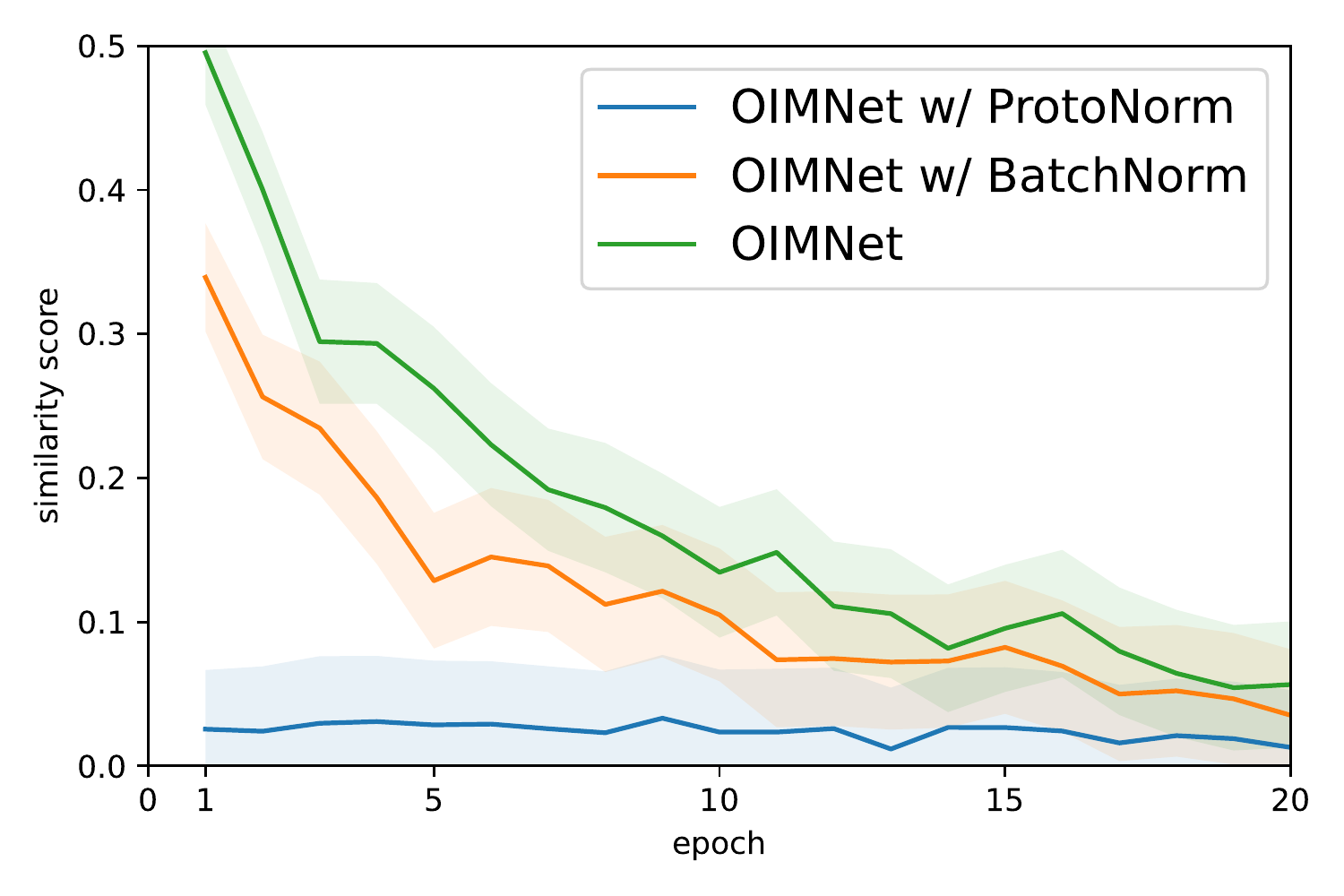}}
  \end{adjustbox}

  \captionsetup{font={small}}
  \vspace{-0.30cm}
  \caption{We plot average cosine similarity scores between features in (a)~the LUT, (b)~the circular queue, and a (c)~concatenation of the two, over training epochs on the PRW dataset~\cite{zheng2017person}. To better demonstrate the advantages of normalization operators, we train all models with the OIM loss. We also illustrate the standard deviation in transparent colors. (Best viewed in color.)}\label{fig:lutcq}
  \vspace{-0.5cm}
\end{figure}

\begin{figure}[t]
  \captionsetup[subfigure]{aboveskip=1pt,belowskip=1pt,justification=centering}

  \begin{adjustbox}{width=0.78\columnwidth,center} % 
      {\includegraphics[width=0.26\columnwidth]{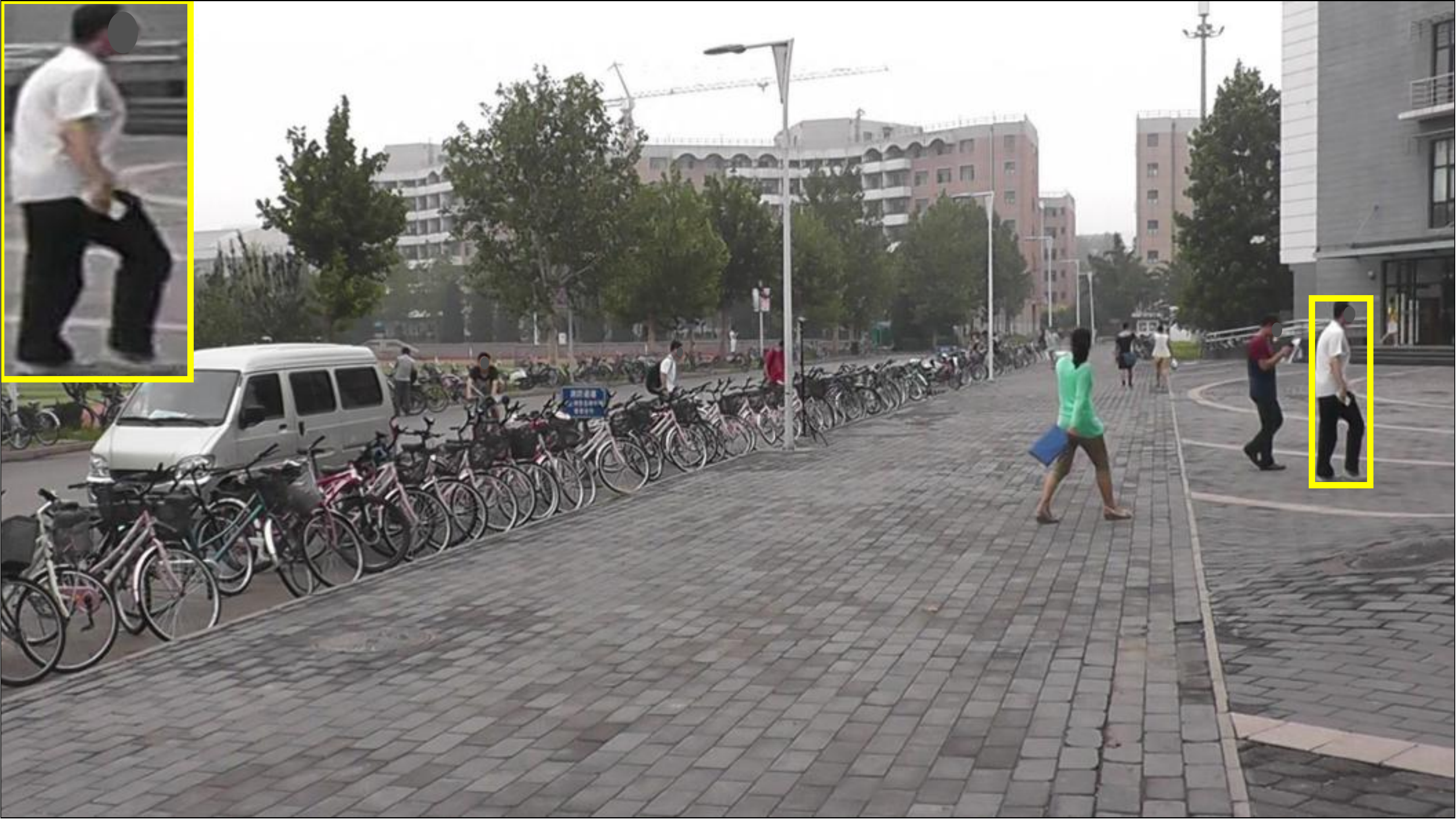}}
      {\includegraphics[width=0.26\columnwidth]{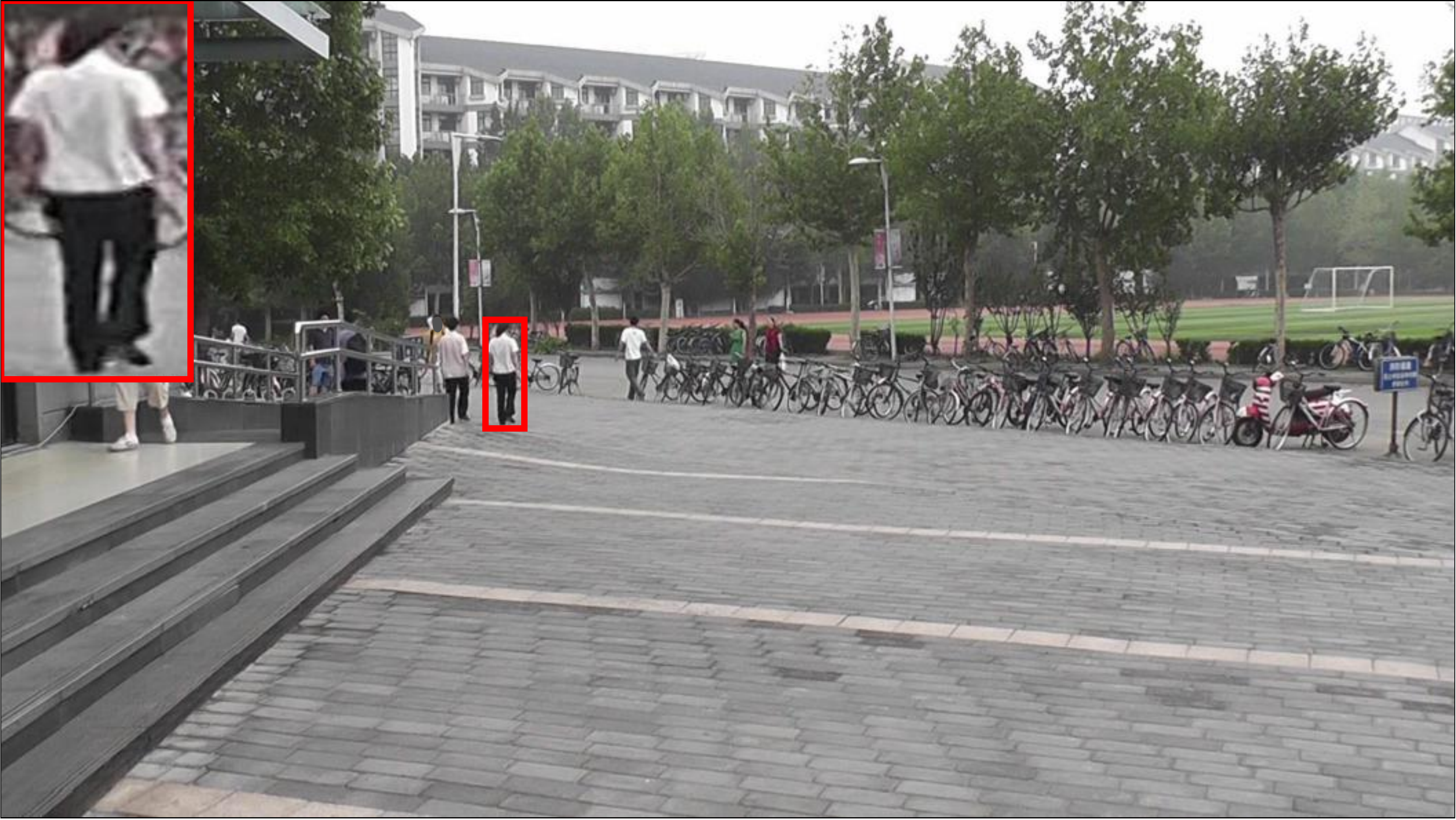}}
      {\includegraphics[width=0.26\columnwidth]{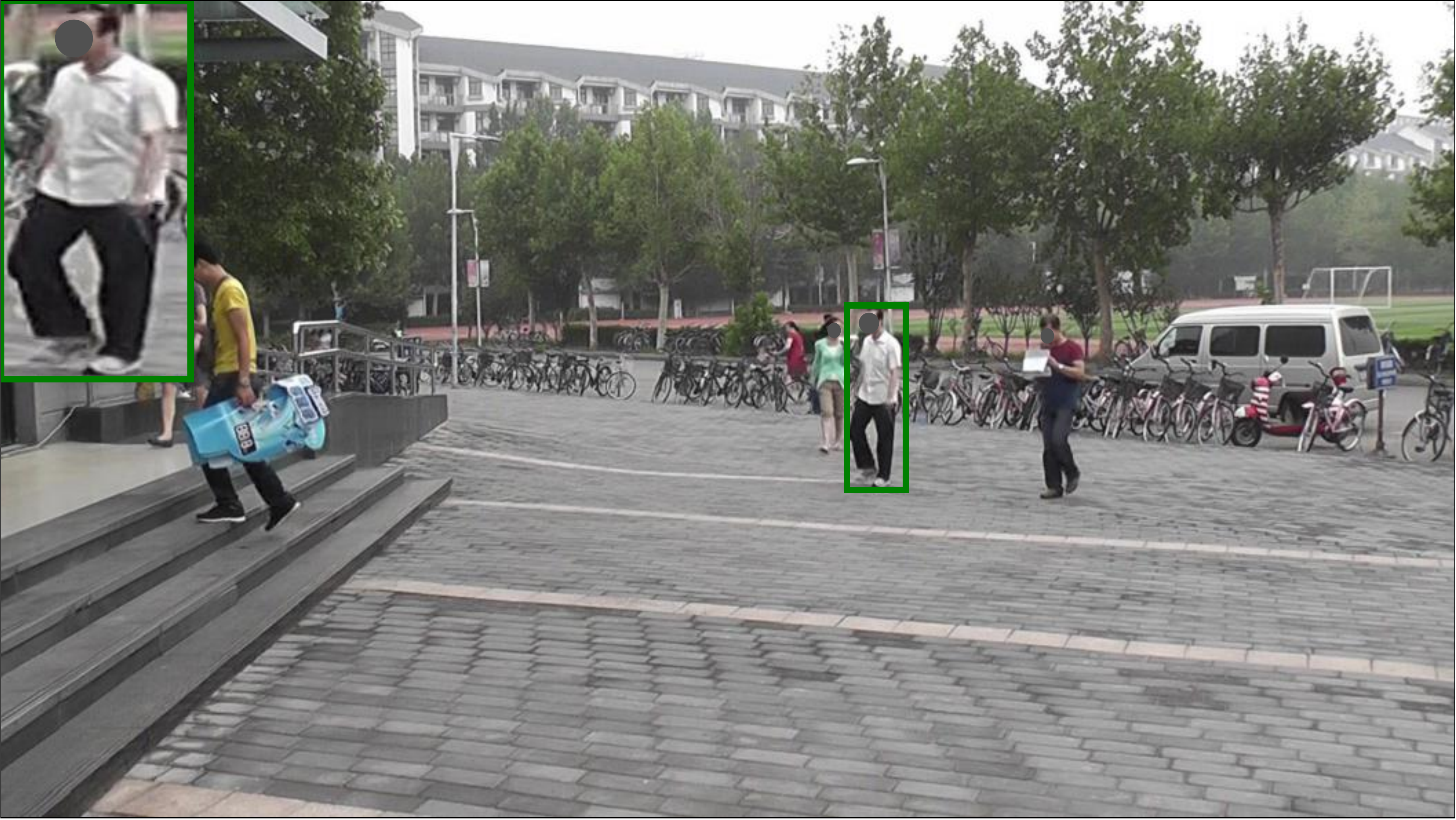}}
  \end{adjustbox}
  
  \begin{adjustbox}{width=0.78\columnwidth,center} % 
      {\includegraphics[width=0.26\textwidth]{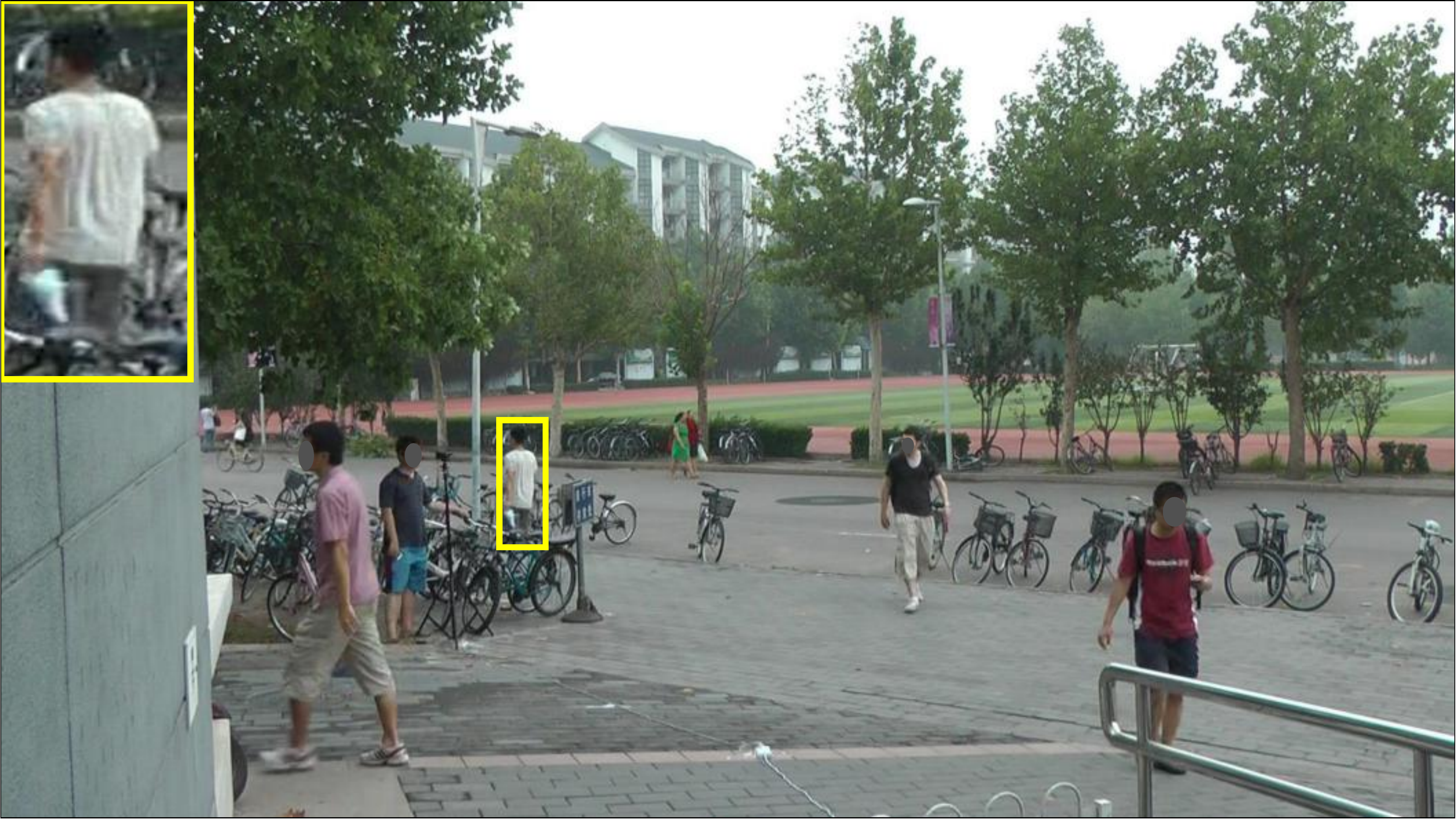}}
      {\includegraphics[width=0.26\textwidth]{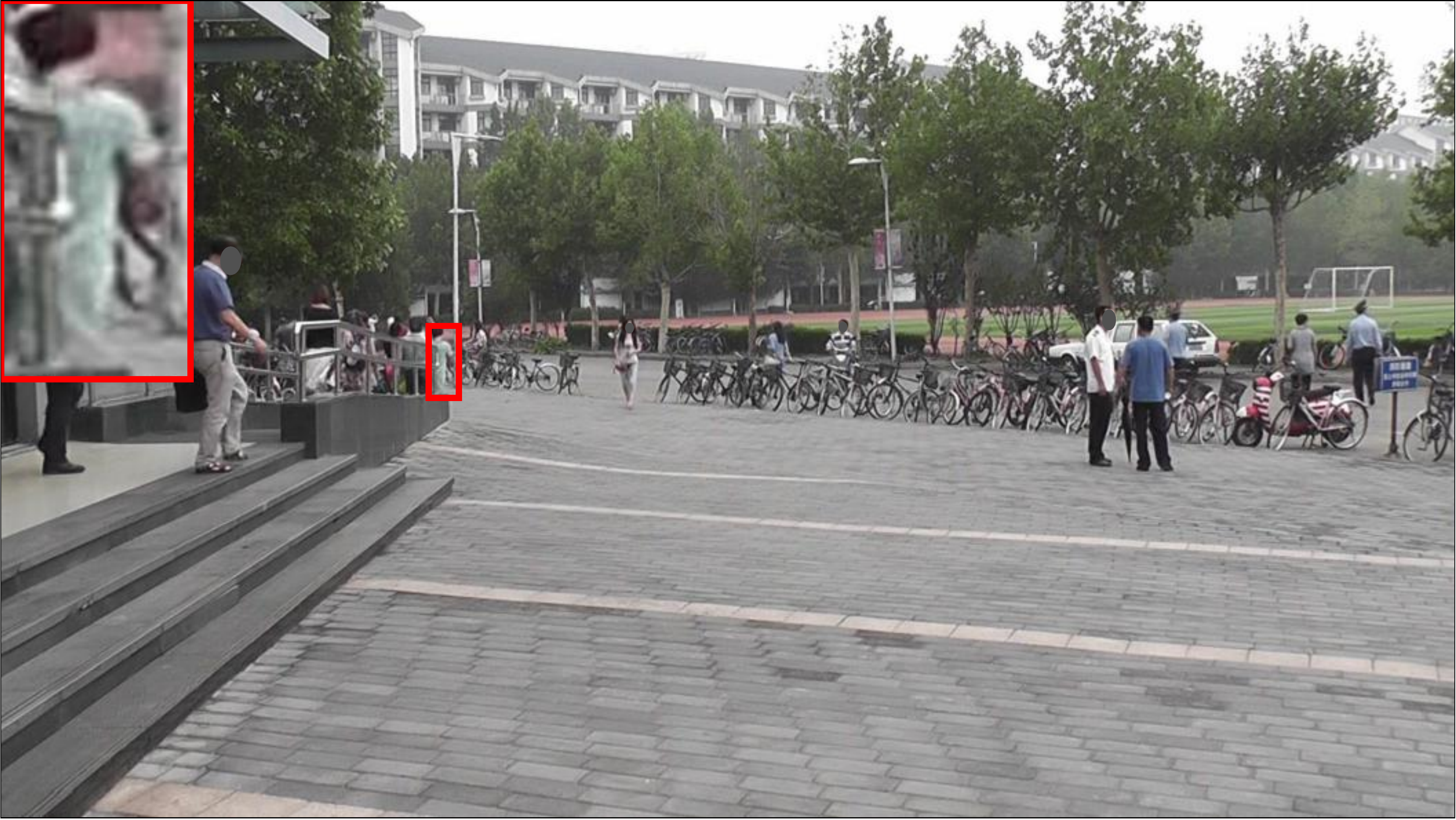}}
      {\includegraphics[width=0.26\textwidth]{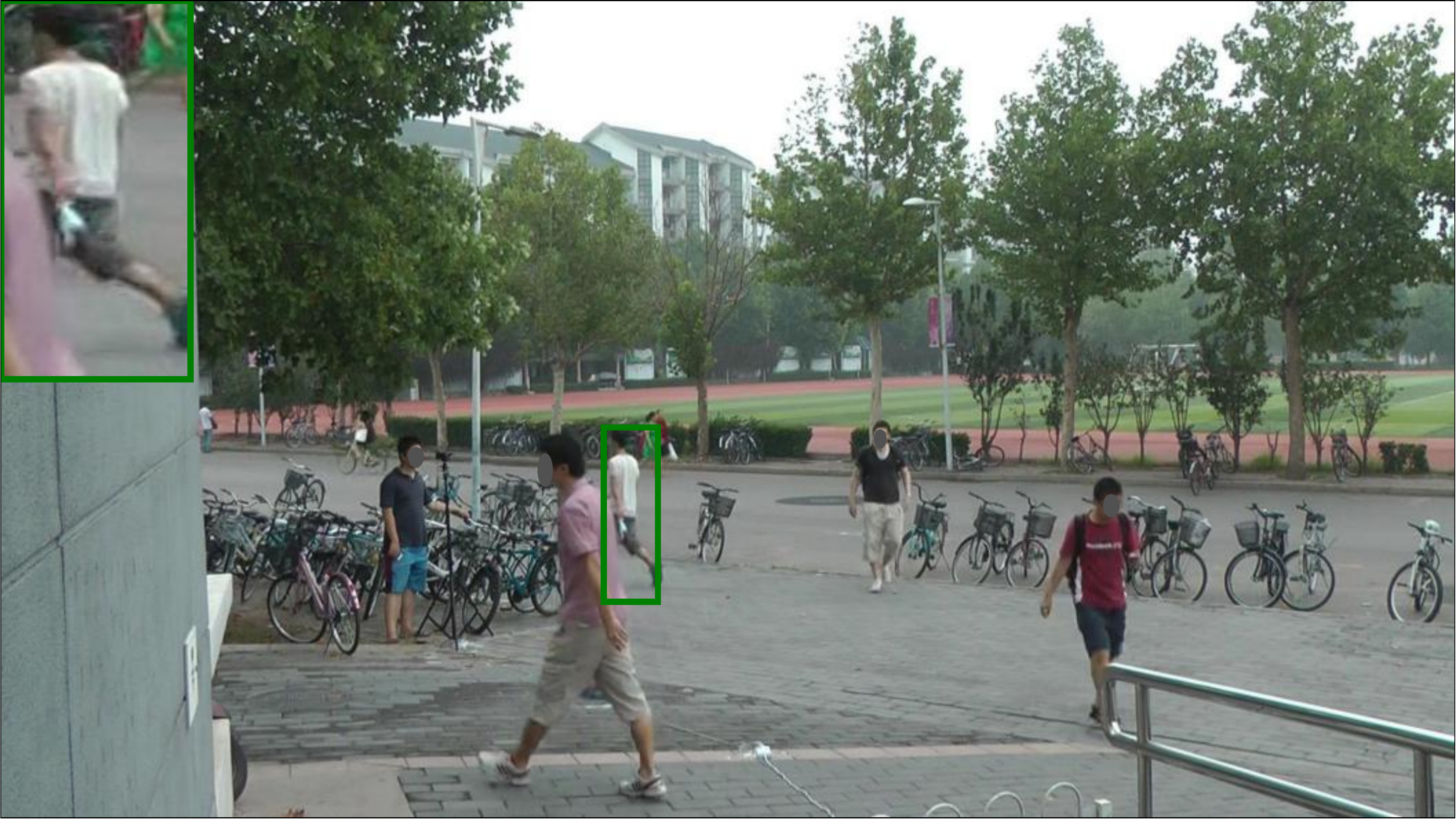}}
  \end{adjustbox}
  
  \begin{adjustbox}{width=0.78\columnwidth,center} % 
      {\includegraphics[width=0.26\textwidth]{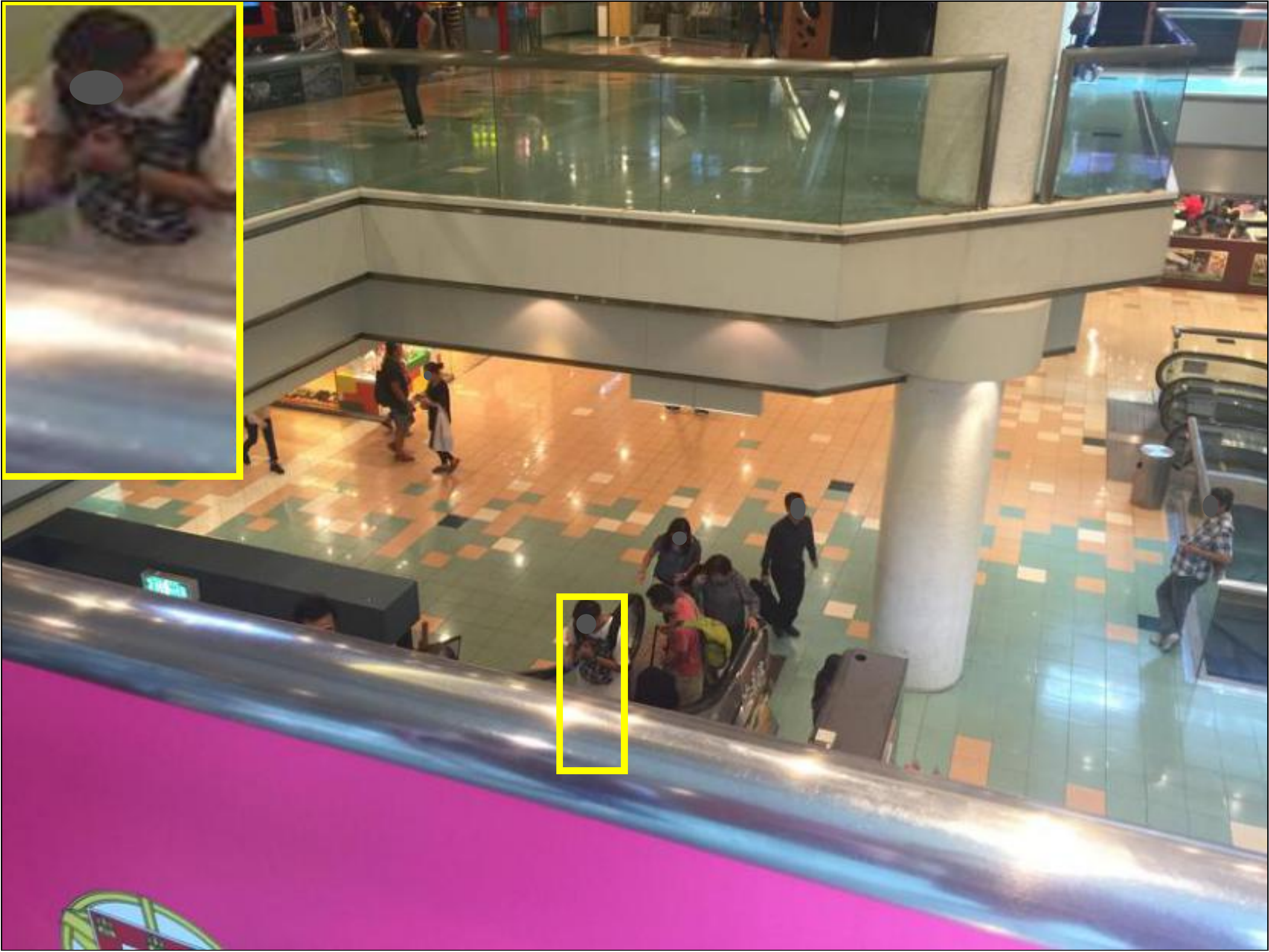}}
      {\includegraphics[width=0.26\textwidth]{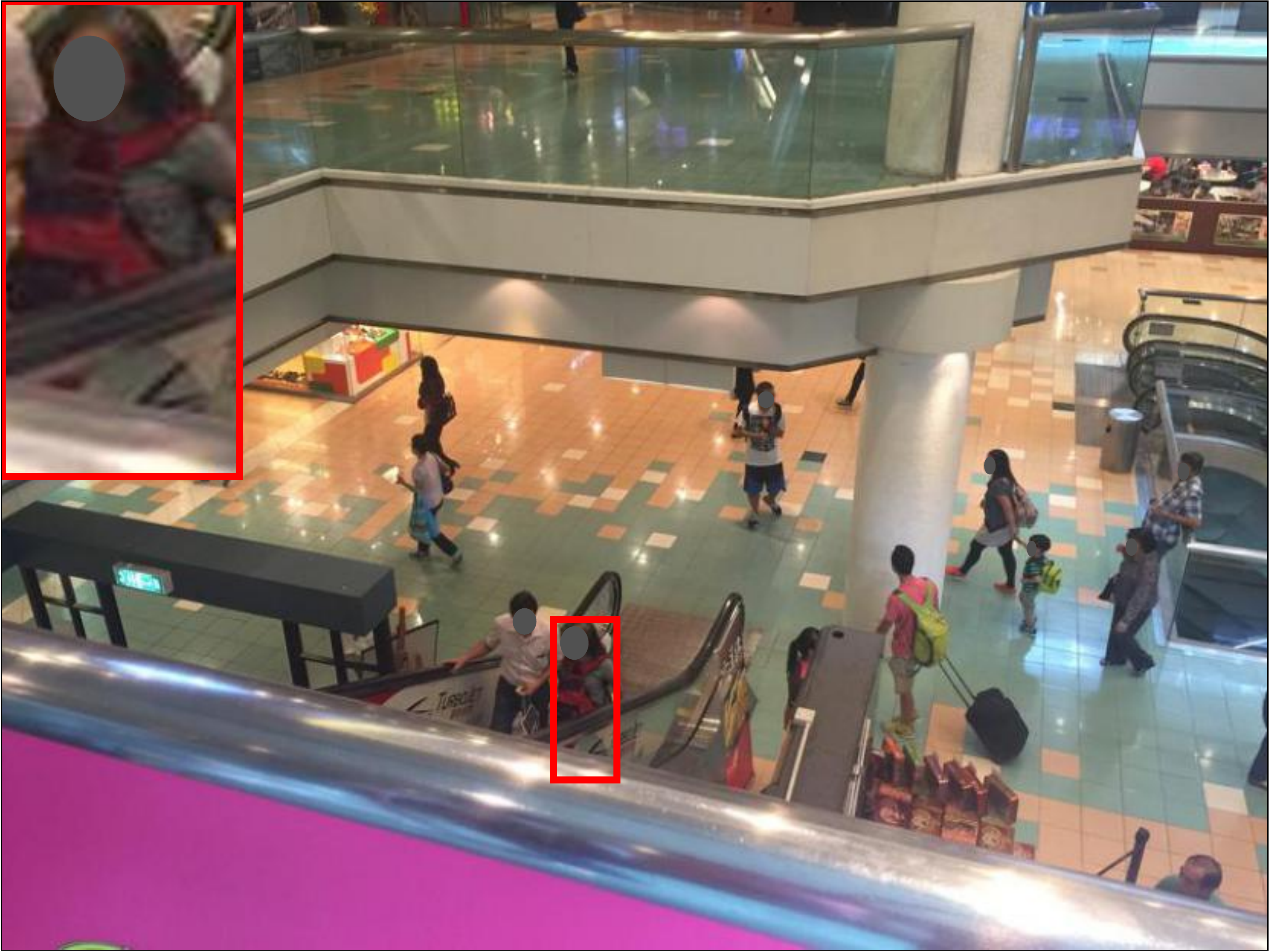}}
      {\includegraphics[width=0.26\textwidth]{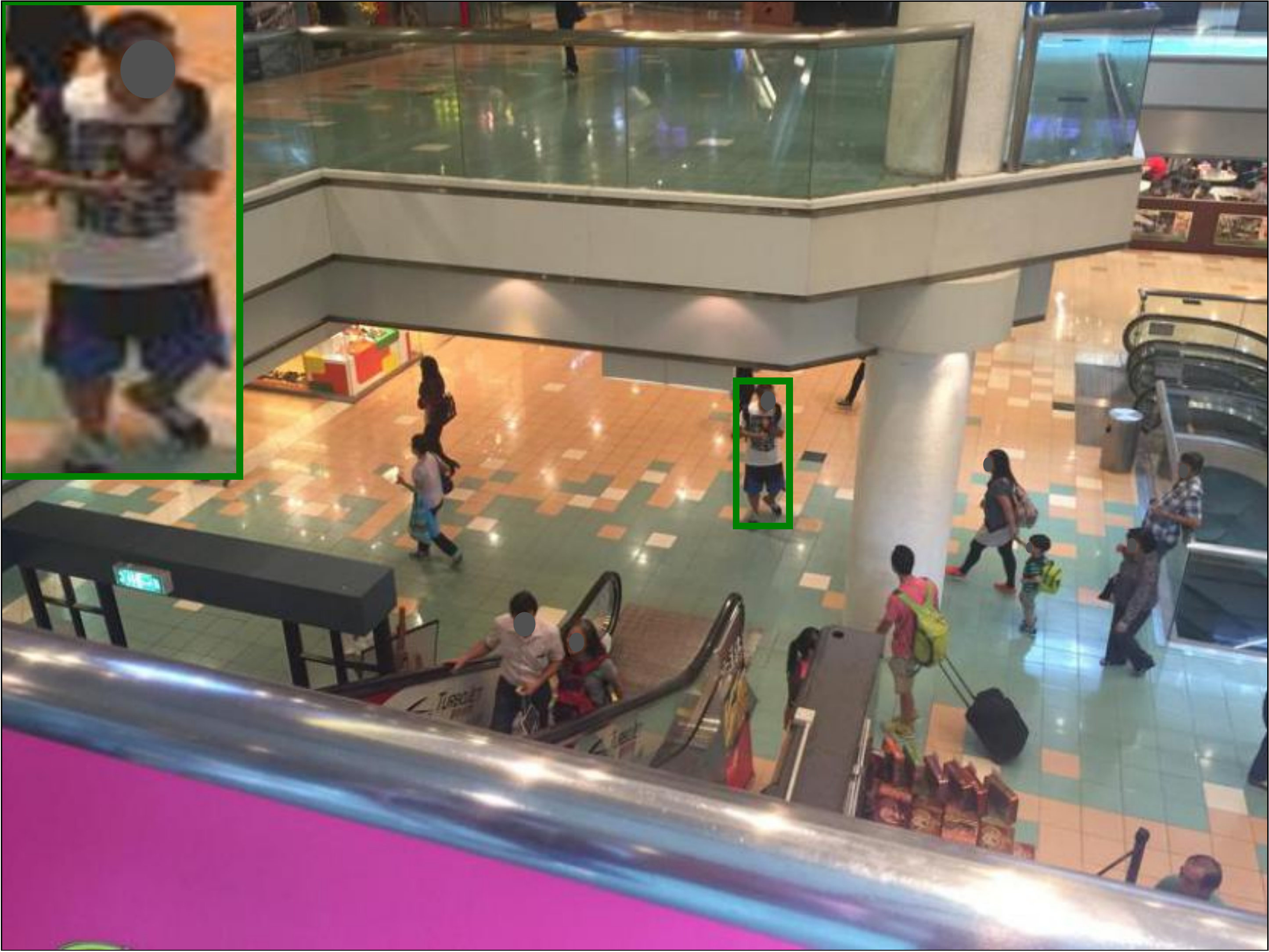}}
  \end{adjustbox}
  
  \begin{adjustbox}{width=0.78\columnwidth,center} % 3x set14 12th
    \subcaptionbox{Query}
      {\includegraphics[width=0.26\textwidth]{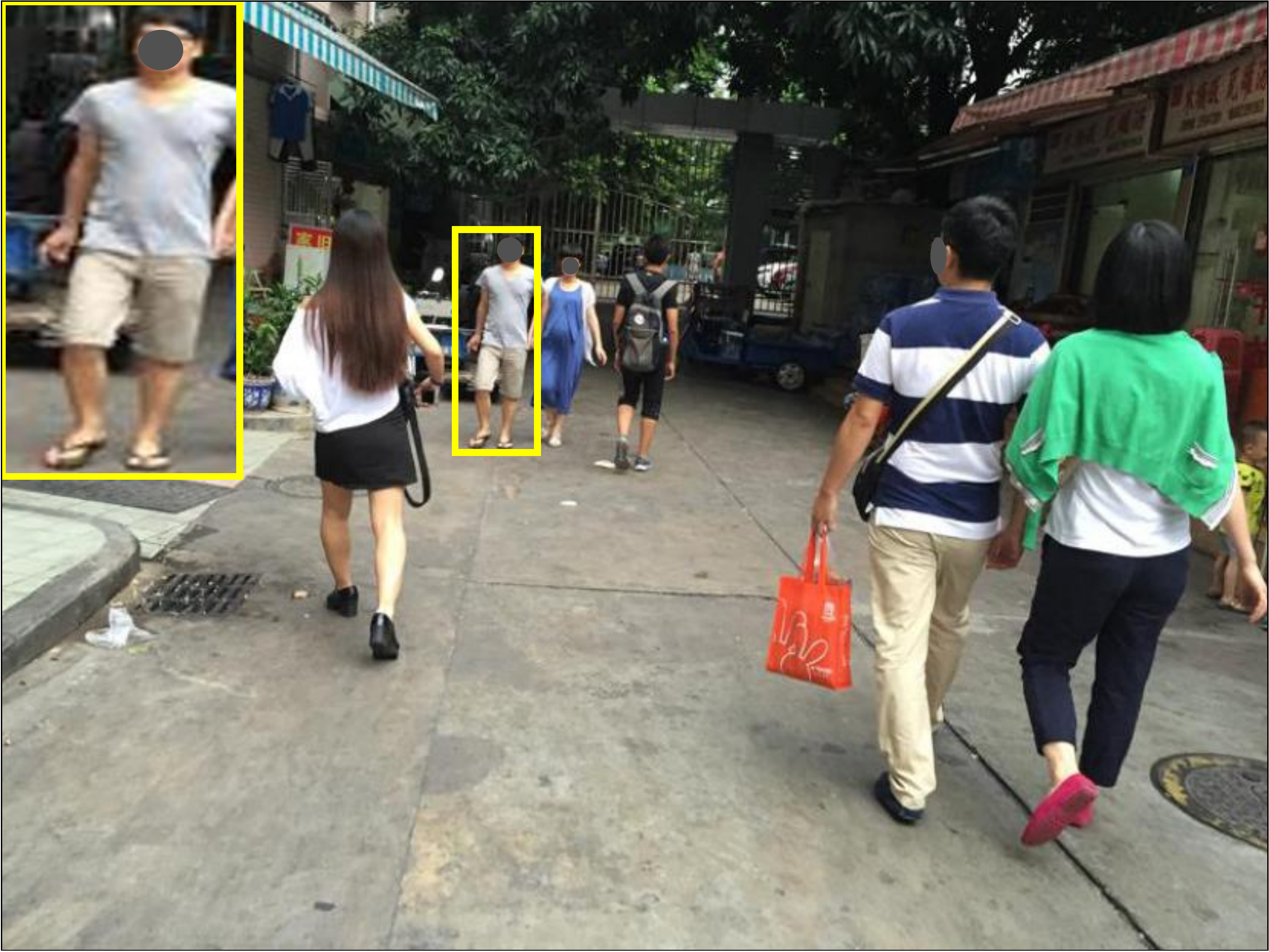}}
    \subcaptionbox{OIMNet~\cite{xiao2017joint}}
      {\includegraphics[width=0.26\textwidth]{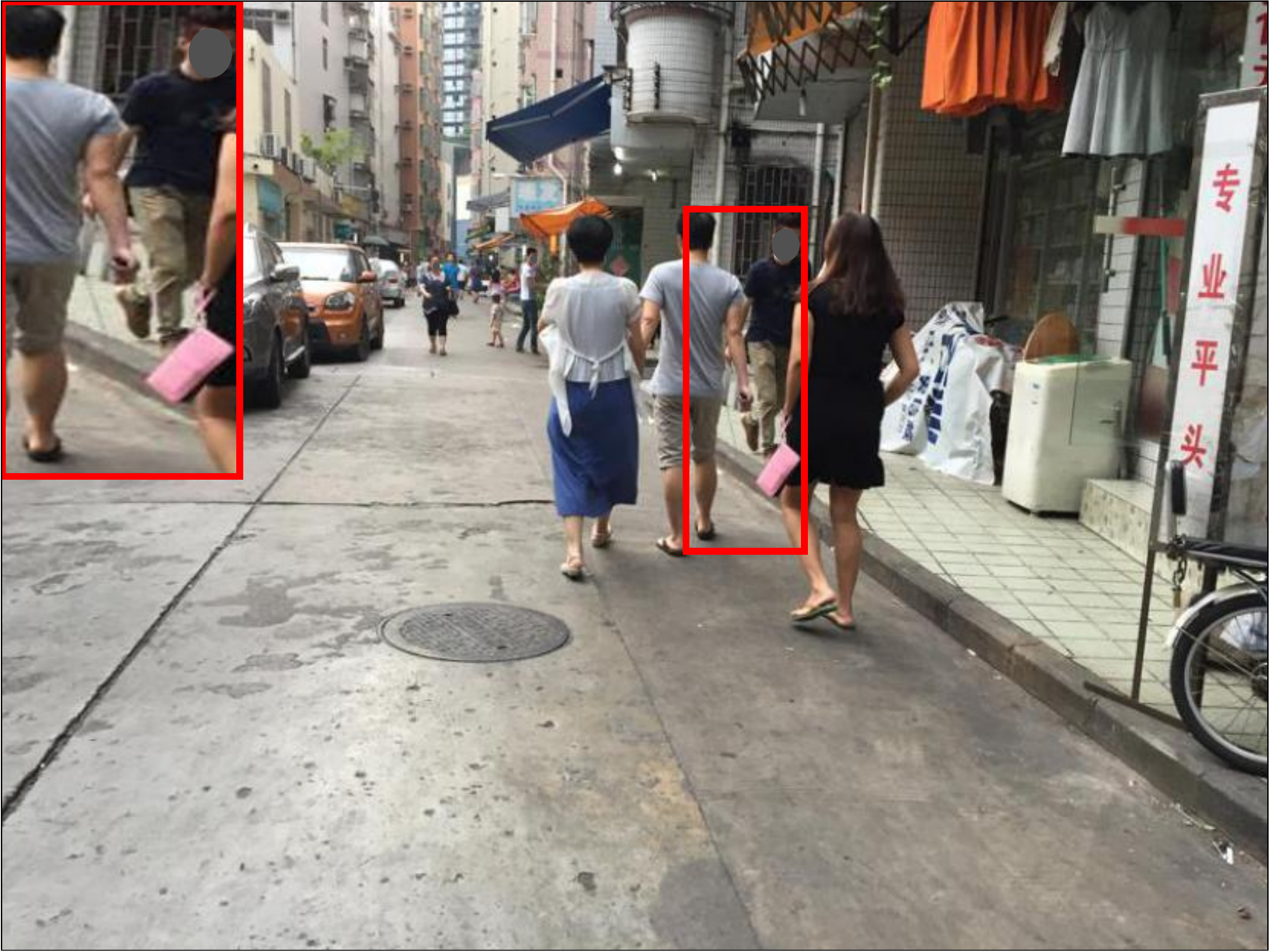}}
    \subcaptionbox{OIMNet\texttt{++}}
      {\includegraphics[width=0.26\textwidth]{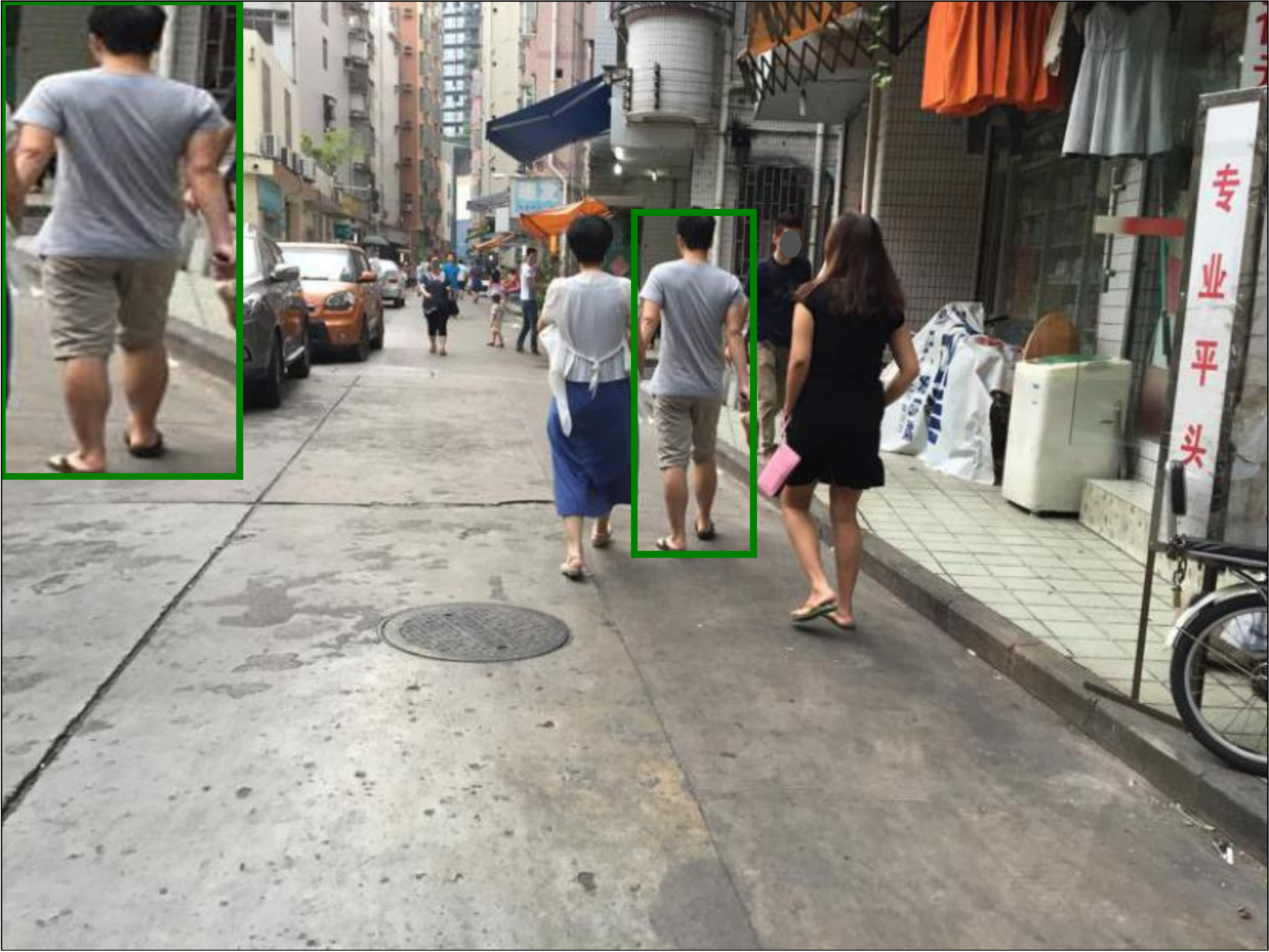}}
  \end{adjustbox}
  
  \vfill
  \vspace{-0.2cm}
  \captionsetup{font={small}}
  \caption{Qualitative comparison between OIMNet~\cite{xiao2017joint} and OIMNet\texttt{++}. For each query image~(left), we visualize top-$1$ search results, where red and green boxes indicate failure and correct cases, respectively. The first two rows are from PRW~\cite{zheng2017person}, and the remaining ones are from CUHK-SYSU~\cite{xiao2017joint}. For each image, we magnify the person-of-interest at the top-left corner for a better visualization.}
  \label{fig:prw}    
\end{figure}

\noindent\textbf{Inter-class separability.} To demonstrate the effectiveness of ProtoNorm, we compute cosine similarity scores between features in the LUT, the circular queue, and both. We average the scores for all possible pairs and show in Fig.~\ref{fig:lutcq} the results over training epochs. We compare the results between three variants of OIMNet~\cite{xiao2017joint}; a vanilla OIMNet, and OIMNets equipped with BatchNorm and ProtoNorm. The variants are trained using the OIM loss. Low similarity scores indicate that the features in the LUT or the circular queue for different IDs encode different information, suggesting a strong inter-class separability. We can observe that employing BatchNorm offers better results in terms of the inter-class separability, compared to the vanilla model, which also demonstrates the importance of calibrating the feature distribution prior to L2 normalizations. We can also see that ProtoNorm obtaining feature statistics less biased towards dominant IDs provides lower similarity scores then BatchNorm, even when trained with a small number of epochs, encouraging more inter-class separability. Moreover, the average distances do not deviate from the initial point severely with ProtoNorm, suggesting that ProtoNorm also stabilizes training process.

%\vspace{-0.15cm}
\noindent\textbf{Qualitative analysis.}
We provide in Fig.~\ref{fig:prw} the visual comparisons between retrieval results for OIMNet~\cite{xiao2017joint} and OIMNet\texttt{++} on PRW~\cite{zheng2017person} and CUHK-SYSU~\cite{xiao2017joint}. We can see that OIMNet\texttt{++} provides person representations that capture subtle discriminative cues, \eg, hair and glasses~(first row), as ProtoNorm in OIMNet\texttt{++} enhances the inter-class separability. We can also observe the effectiveness of the LOIM loss. For example, OIMNet\texttt{++} is more robust to occlusions~(second and third row) and person overlaps~(fourth row), since the LOIM loss favors pedestrian proposals with better localization accuracies to train with features in the LUT.\\

\section{Conclusion}
We have introduced OIMNet\texttt{++} for person search that addresses the limitations of existing methods. To this end, we have presented a novel normalization scheme, dubbed ProtoNorm, that provides better statistics for feature standardization, even under the extreme class imbalance across person IDs. We have also introduced the LOIM loss that exploits the localization accuracy of each proposal to learn more discriminative representations. Finally, we have demonstrated the effectiveness of each component with extensive ablation studies, and have shown that OIMNet\texttt{++} outperforms other person search methods on the standard person search benchmarks by a large margin.\\

\noindent\footnotesize{\textbf{Acknowledgements.} This work was partly supported by Institute of Information $\&$ communications Technology Planning $\&$ Evaluation~(IITP) grant funded by the Korea government~(MSIT)~(No.RS-2022-00143524, Development of Fundamental Technology and Integrated Solution for Next-Generation Automatic Artificial Intelligence System, and No.2021-0-02068, Artificial Intelligence Innovation Hub), the Yonsei Signature Research Cluster Program of 2022~(2022-22-0002), and the KIST Institutional Program~(Project No.2E31051-21-203).}

\clearpage
\bibliographystyle{splncs04}
\bibliography{egbib}

\clearpage


\section*{Supplementary material}

# OIMNet++: Prototypical Normalization and Localization-aware Learning for Person Search

Sanghoon Lee[1], Youngmin Oh[1], Donghyeon Baek[1],
Junghyup Lee[1], and Bumsub Ham[1,2]★

[1]Yonsei University    [2]Korea Institute of Science and Technology (KIST)

In this supplementary material, we present qualitative analysis on head/tail classes for person search (Sec. S1) and visualizations of positive proposals that sufficiently overlap with ground-truth bounding boxes during training (Sec. S2).

## S1    Quantitative analysis

ProtoNorm calibrates feature distributions while considering minority IDs during training. To demonstrate the effectiveness, we categorize training IDs into head and tail classes by splitting at the median occurrence value. For each ID, we compute cosine similarity scores with all features within the LUT. We then average the scores w.r.t the head/tail classes separately and visualize in Fig. S1 the results. Lower similarity scores indicate stronger inter-class separability, indicating higher discriminative power of person representations. We compare the results between three variants; vanilla OIMNet [2], OIMNets adopting either BatchNorm [1] or ProtoNorm. We can observe that vanilla OIMNet shows the worst inter-class separability as it does not consider feature distribution prior to L2 normalization. Employing BatchNorm alleviates this problem. This, however, calibrates feature distribution without considering the long-tail distribution across person IDs, restraining the discriminative power of person representations. ProtoNorm computes feature statistics that are less biased to head IDs for the calibration. This allows to encourage inter-class separateness regardless of the sample distribution across person IDs. As a result, ProtoNorm strengthens the inter-class separability offers lowest similarity scores for both cases.

## S2    Visualizations of positive proposals

During training, we have observed that features extracted from positive proposals contain different extents of noise, *e.g.*, background clutter and person overlaps. We provide in Figs. S2 and S3 the visual examples of positive proposals on the PRW [3] and CUHK-SYSU [2] datasets, respectively. These proposals overlap with the corresponding ground-truth bounding boxes more than 0.5 intersection-of-union (IoU) scores. Nevertheless, they depict detrimental noise that can distract learning discriminative person representations. For both figures, proposals in the left three columns contain background clutter, and the remaining ones depict other pedestrian instances. The LOIM loss assigns adaptive momentum value for each proposal, discouraging severely misaligned proposals to contribute in discriminative feature learning for re-identification.

---

★ Corresponding author.

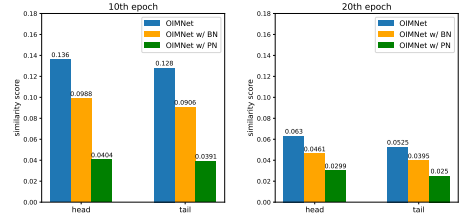

Fig. S1: Average cosine similarity scores on PRW [3] for variants of OIMNet [2]. We divide person IDs into head and tail classes and average the scores among each category over training epochs. Lower similarity score indicates stronger inter-class separateness.

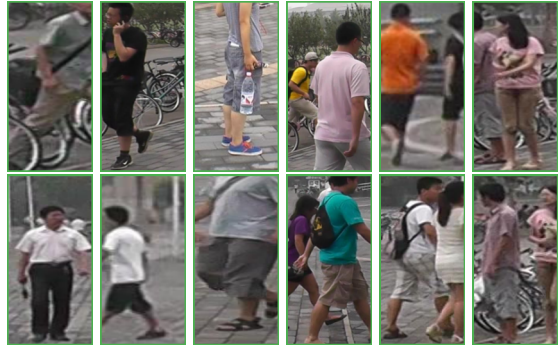

Fig. S2: Positive proposals obtained from PRW [3]. Examples in the top row are sampled at the 10-th epoch, while the bottom ones are sampled at the 20-th epoch.

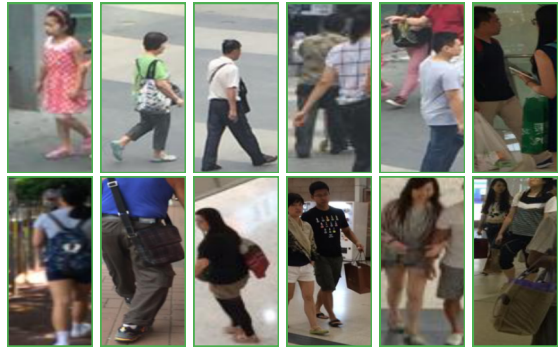

Fig. S3: Positive proposals obtained from CUHK-SYSU [2]. Examples in the top row are sampled at the 10-th epoch, while the bottom ones are sampled at the 20-th epoch.


\end{document}